%% file: paper.tex
\documentclass[journal]{IEEEtran}

\input{tex/preamble}
\ifCLASSINFOpdf
\else
\fi
\begin{document}
\ifdraft{\listoftodos}{}%
%
\title{Robotic Picking of Tangle-prone Granular Materials using Parallel Grippers}
%
%
%

\author{Prabhakar Ray,~\IEEEmembership{Student Member,~IEEE,}
        Matthew Howard,~\IEEEmembership{Member,~IEEE}
\thanks{P. Ray and M. Howard are with the Centre for Robotics Research, Department of Engineering, King's College London, London, UK. \texttt{\{ray.prabhakar,matthew.j.howard\}@kcl.ac.uk}}
}

%
%

\markboth{IEEE TRANSACTIONS ON ROBOTICS, VOL. X, NO. X, XXX 20XX}%
{Shell \MakeLowercase{\textit{et al.}}: Bare Demo of IEEEtran.cls for IEEE Journals}
%



\maketitle

\begin{abstract}
The picking of one or more objects from an unsorted pile continues to be non-trivial for robotic systems. This is especially so when the pile consists of a \gls{GM} containing individual items that tangle with one another, causing more to be picked out than desired. One of the key features of such \emph{tangle-prone \glspl{GM}} is the presence of \emph{protrusions} extending out from the main body of items in the pile. This work characterises the role the latter play in causing mechanical entanglement and their impact on picking consistency. It reports experiments in which picking \glspl{GM} with different \PLs results in up to $76\%$ increase in picked mass variance, suggesting \PL to be an informative feature in the design of picking strategies. Moreover, to counter this effect, it proposes a new \SP approach that significantly reduces tangling, making picking more consistent. Compared to prior approaches that seek to pick from a tangle-free point in the pile, the proposed method results in a decrease in picking error (\e) of up to $51\%$, and shows good generalisation to previously unseen \GMs.
\end{abstract}

\begin{IEEEkeywords}
Robotics in Agriculture and Forestry, Agricultural Automation, Computer Vision for Automation
\end{IEEEkeywords}

%
\IEEEpeerreviewmaketitle

\section{Introduction}
%
%
%
%
\IEEEPARstart{T}HE automated picking of one or more objects from an unsorted pile or bin is a common task in many manufacturing processes. One of the primary challenges to robotic automation of this task is dealing with the mechanical entanglement that inevitably occurs between different objects in the bin/pile, causing undesired items to be picked along with the target object(s). 

Much attention has been given to solving this issue, especially for large objects (see \fref{intro_pic}\ref{f:non_gm_obj}), for instance, by planning pick operations that avoid tangled objects \cite{Matsumura2019LearningObjects} or through physical pile interaction such as pushing \cite{Kaipa2016ResolvingBinPicking}. However, few studies consider the robotic handling of more challenging granular materials (\glspl{GM}) consisting of a collected mass of small items, such as occurs when handling edible materials like grains or salads (see \fref{intro_pic}\ref{f:gm_non_tangle_obj} and \ref{f:gm_tangle_obj_herbs}). For the latter, usually the aim is to accurately extract a smaller specified quantity from a larger mass, \emph{the difficulty of which depends on whether the \GM tangles or not}. The extraction of quantities of \emph{non-tangling} \glspl{GM} has received some attention, for instance \textcite{Schenck2017LearningMedia} explore the manipulation of pinto beans. However, the issue of \emph{picking excess mass due to entanglement} such as occurs in \emph{severely tangle-prone \glspl{GM}}, remains largely unexplored. This work is the first to \il{\item characterise the propensity of a \GM to tangle in terms of a measurable quantity, and \item propose strategies to mitigate the effect of tangling in robotic picking.}
\begin{figure}[t!]
\centering
\begin{tikzpicture}
\node [draw=none] at (-9.3,0.975) {\includegraphics[height=.080\textheight]{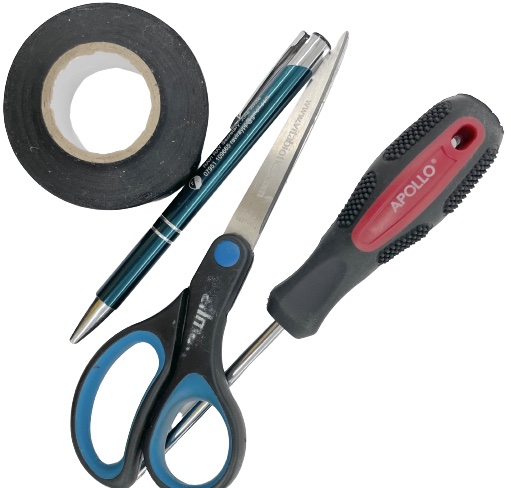} };
\node [draw=none] at (-6.8,0.975) {\includegraphics[height=.080\textheight]{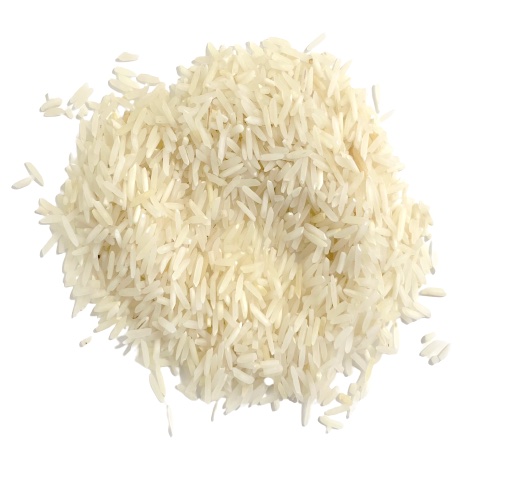} };
\node [draw=none] at (-9.3,-1.47) {\includegraphics[height=.080\textheight]{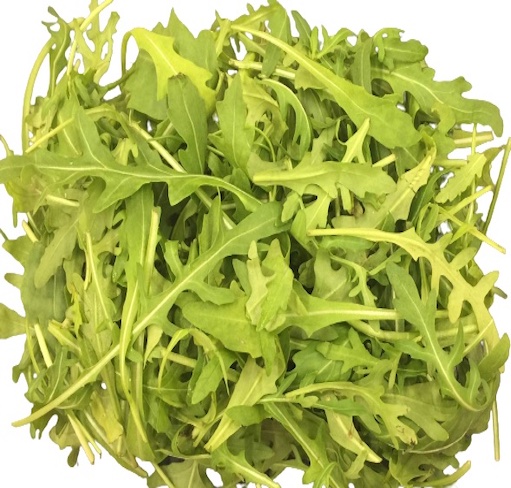} };
\node [draw=none] at (-6.8,-1.47) {\includegraphics[height=.080\textheight]{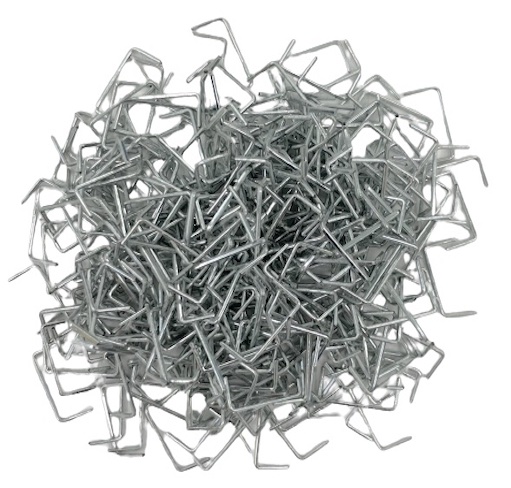} };
\node [draw=none] at (-3.5,-0.25) {\includegraphics[height=.180\textheight]{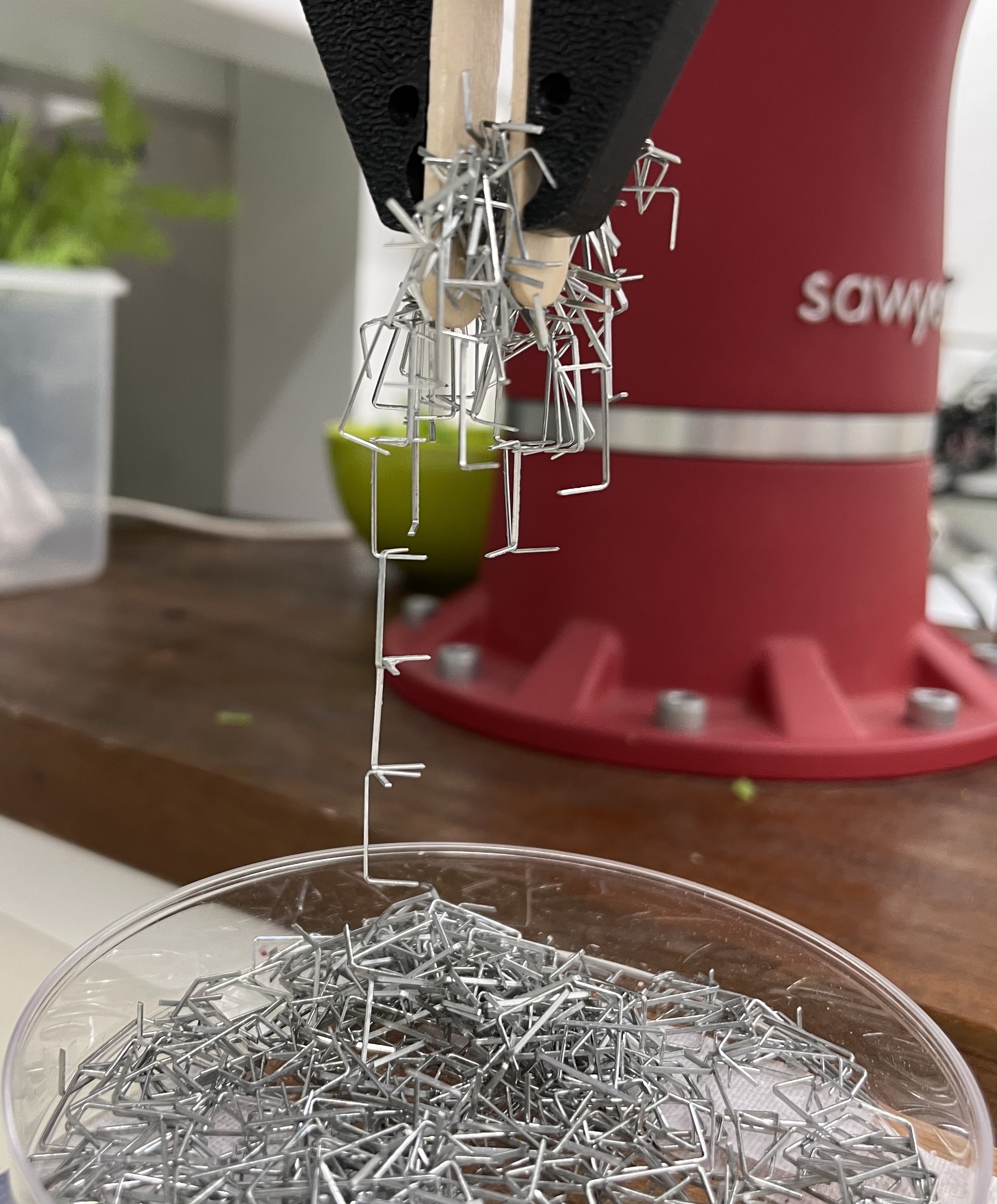} };
\node at (-8.47,0.22){\colorbox{white}{\ref{f:non_gm_obj}}};
\node at (-5.99,0.22){\colorbox{white}{\ref{f:gm_non_tangle_obj}}};
\node at (-8.47,-2.2){\colorbox{white}{\ref{f:gm_tangle_obj_herbs}}};
\node at (-5.99,-2.2){\colorbox{white}{\ref{f:gm_tangle_obj_staples}}};
\node at (-1.95,-2.22){\colorbox{white}{\ref{f:picked_staples_intro}}};
\end{tikzpicture}
	\caption{
	\cl{%
	\item\label{f:non_gm_obj}Traditional bin composed of non-granular (large) objects.%
	\item\label{f:gm_non_tangle_obj} A granular, non-tangling material (rice grains). Examples of tangle-prone \GMs include%
	\item\label{f:gm_tangle_obj_herbs} herbs (wild rocket) and  %
	\item\label{f:gm_tangle_obj_staples} staples. %
\item\label{f:picked_staples_intro} Tangling makes the mass lifted in a simple pick operation difficult to predict.%
			}%
		}%
	\label{f:intro_pic}
\end{figure}
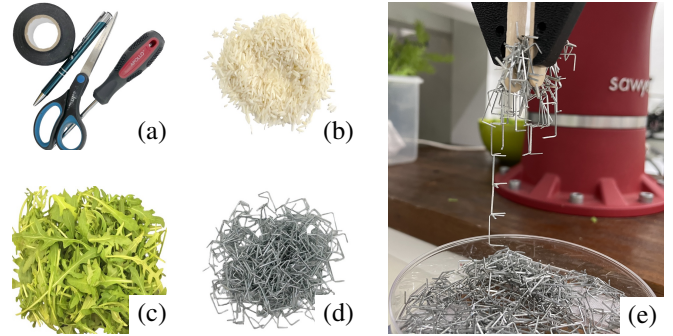

Specifically, it evaluates the \PL of a \GM as an indicator of tangling propensity that can be used to predict picking consistency when attempting one-shot picking under a pre-specified constraint, such as desired mass or number of items to be picked. Moreover, it proposes \acrfull{SnP} as a method to deal with the inherent uncertainty introduced by tangling (see \fref{intro_pic}\ref{f:picked_staples_intro}).
The effectiveness of this method in improving picking consistency by reducing tangling is evaluated in two experiments with several different \glspl{GM}, extending the work presented by \textcite{RayHerbUntangling}. 

In experiments where a $7$-\gls{DoF} robot with a parallel gripper is used to pick pre-set quantities from tangled piles of staples and herbs, a significant increase ($76\%$) in the picked mass variance is observed for \GMs with non-zero \PL, suggesting protrusions play an important role in causing tangling and making picking inconsistent. Compared to the approach of avoiding tangling by seeking to pick from a tangle-free point in the pile, the proposed method results in a decrease in \e\ of up to $51\%$, and shows good generalisation to previously unseen \GMs. This highlights the benefit of using \gls{SnP} as a practical tool for deploying robotic automation for a variety of challenging picking tasks involving \glspl{GM}, such as mass-constrained herb packaging.


\section{Related Work}\label{s:related_work}
%
%
%

The picking of objects from containers is frequently termed the \emph{bin-picking problem} and has a long history in the robotic automation literature.

Bins composed of large objects (see \fref{intro_pic}\ref{f:non_gm_obj}) have received much more attention than \glspl{GM}. \textcite{Taylor2003RobustGeometry1} propose using simple geometric primitives such as planes, spheres, cylinders and cones for object recognition in the bin. Changes in surface types and depth discontinuities are then used to segment the cluttered scene. A vision-based algorithm is proposed in \cite{Xu2015RoboticTray} to resolve gripper-object collision by identifying and picking the topmost object in a pile composed of surgical instruments. \textcite{Schwarz2017PointNetPicking} propose a deep learning approach for extracting large individual objects from a cluttered bin. \textcite{push_grasp_syn} propose learning synergies between pushing and grasping to improve grasping success rates. These methods consider bins composed of large objects and prove effective for avoiding gripper-object collision but not object entanglement.

The issue of object entanglement has received some attention for bins composed of large objects such as industrial parts. \textcite{KaipaAutomatedBins.pdf} use CAD models for estimating a singulation plan for tangle-free extraction of individual objects from a heterogeneous tangle-prone pile. Singulation plans encountering object entanglement are discarded. A human-robot collaboration approach is proposed in \cite{Kaipa2014Human-robotAssemblies} for resolving grasping errors due to issues such as occlusion and random object postures, including entanglement. 
\textcite{AvoidingEntanglement} propose a method for increasing the robustness of bin picking by avoiding grasps of entangled objects. Although the methods here consider tangling directly, their objective is to extract a single individual object by avoiding entangled scenarios. In contrast, this work considers extracting a uniform quantity of \glspl{GM} consistently, especially when tangling cannot be avoided.

Non-tangling \glspl{GM} (see \fref{intro_pic}\ref{f:gm_non_tangle_obj}) have been studied in the context of robotic \il{\item scooping \cite{scoop_gm1,scoop_gm2} and \item pouring \cite{pouring_gm1,pouring_gm2}.} \textcite{Kuriyama2019AMaterials} present a soft pneumatic gripper for packaging non-tangling food materials such as kernel corn. The authors report that although the amount (mass) of material picked using the gripper can be controlled by varying the insertion depth, the variation among trials is significant---due to the bending of the soft gripper material. 

Robotic picking of tangle-prone \glspl{GM} (see \fref{intro_pic}\ref{f:gm_tangle_obj_herbs} and \ref{f:gm_tangle_obj_staples}) under external constraints such as mass or number of items remains largely unexplored. 
In terms of objective, perhaps the closest work to the present study is that of \textcite{EntangledFoodGripper} where a pre-grasping motion is proposed for countering issues such as  adhesion and object entanglement in bins composed of food materials such as shredded cabbage and bean sprouts. The pre-grasping motion consists of a sequence of actions where the food is picked up and dropped before repeating the pick from the same point. However, the work presented here specifically focuses on entanglement reduction through a separation strategy without having to repeat the pick. 


\glspl{GM} have also been studied in the context of entanglement and pile stability. \textcite{Barabasi_gm1} propose stability criteria for calculating the maximum angle of stability for homogeneous \glspl{GM}  composed of 3D spherical particles and 2D circular discs. \textcite{Bocquet_gm2} explore the relationship between cohesion forces and maximum avalanche angle for rough spherical beads. Penetration studies involving soil and sand also provide valuable insights into the physical dynamics of \glspl{GM} \cite{gm3}. However, most studies involve approximately spherical (convex) \glspl{GM}, and the shape of the particles has not received much attention \cite{gm4}.

To date, the issue of \emph{picking excess mass due to entanglement} such as occurs in \emph{tangle-prone \glspl{GM}}, has not received much attention. This work is the first to \il{\item characterise the propensity of a \GM to tangle in terms of a measurable quantity, and \item present strategies to mitigate the effect of tangling} to achieve a level of predictability in robotic picking.










\section{Problem Definition}\label{s:problem_def}
This work considers the problem of picking a target mass from a pile of tangle-prone \glspl{GM} such as L-hooks, cup hooks, staples and herbs.
In the context of mass-constrained bin picking, the primary objective of the robot is to pick a target mass accurately. The picking error \e\ is expressed as
\begin{equation}
\e = \frac{1}{N} \sum_{n=1}^{N} |\mt-\m_n|,
\label{e:perr_eq}
\end{equation}
where $\mt$ is the target mass and $\m_n$ is the picked mass for trial $n$. %
The objective is to learn to pick in a way that minimises \eref{perr_eq} for any given $\mt$. The desired picking skill expressed as
\begin{equation}
\pickpar = \f(\mt),
\label{e:pickpar_eq}
\end{equation}
where $\f(\mt)$ maps the target mass $\mt \in {\displaystyle \mathbb{R} _{>0}=\left\{x\in \mathbb{R} \mid x>0\right\}}$ to pick parameter $\pickpar = (\br,\w)^\T$ comprising of a picking location $\br=(\rx,\ry,\rt)^\T$ with gripper orientation $\rt$ around the vertical ($z$) axis and gripper aperture $\w$, enables the selection of $\pickpar$ such that \eref{perr_eq} is minimised for the target mass $\mt$. 
However, the highly stochastic nature of tangle-prone pile of \glspl{GM} makes such minimisation non-trivial. For example, for a fixed pile mass, container volume and pick parameter $\pickpar$ as estimated for a target mass $\mt$, a consistent mass ($m_1^{\pickpar} = m_n^{\pickpar}$) is expected to be picked across trials but instead the pile entanglement leads to a high \e. 
Considering the simplest case of picking using a fixed picking parameter $\pickpar$ for a target mass $\mt$, \eref{perr_eq} can be reduced simply by adjusting the gripper aperture $\w$ based on the degree of pile entanglement. However, estimating the degree of entanglement in a \gls{GM} pile is non-trivial. Additionally, to improve the consistency and predictability of picking, \e\ variance arising out of pile entanglement should be reduced as far as possible.
A lower degree of pile entanglement will reduce \e\ variance, making picking more predictable. To this end, this work proposes a separation strategy (\SP) to effectively reduce \e\ without directly estimating the degree of entanglement in a pile for efficient mass-constrained robotic bin-picking for bins composed of \glspl{GM} such as staples and herbs. 

%

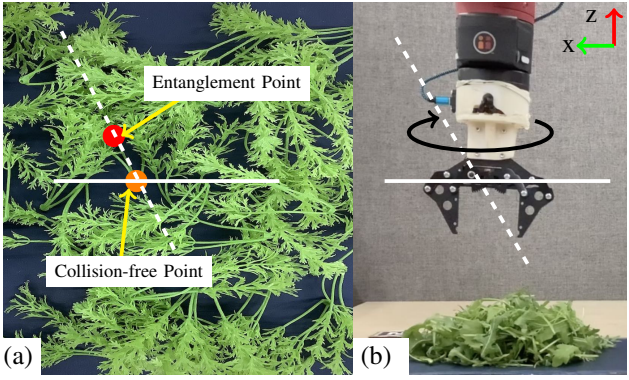
\begin{figure}[t!]%
\centering%
\begin{tikzpicture}%
\def\E{1.2}
\node at (-6.9,0.60){\includegraphics[width=.28290\textwidth]{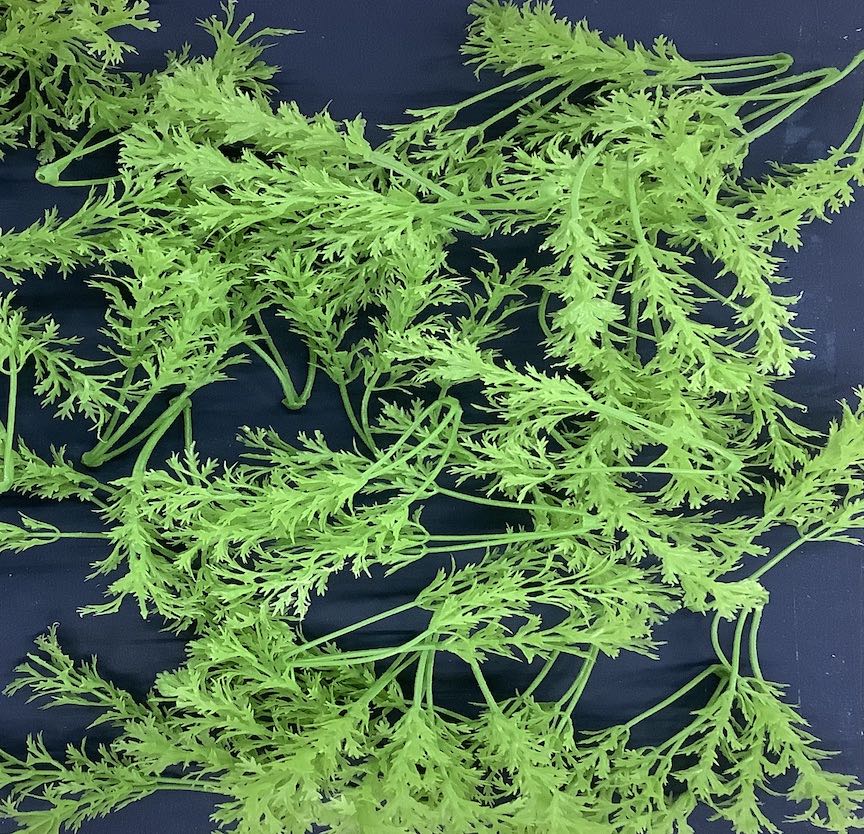}};%
\node at (-3,0.60){\includegraphics[width=.2\textwidth]{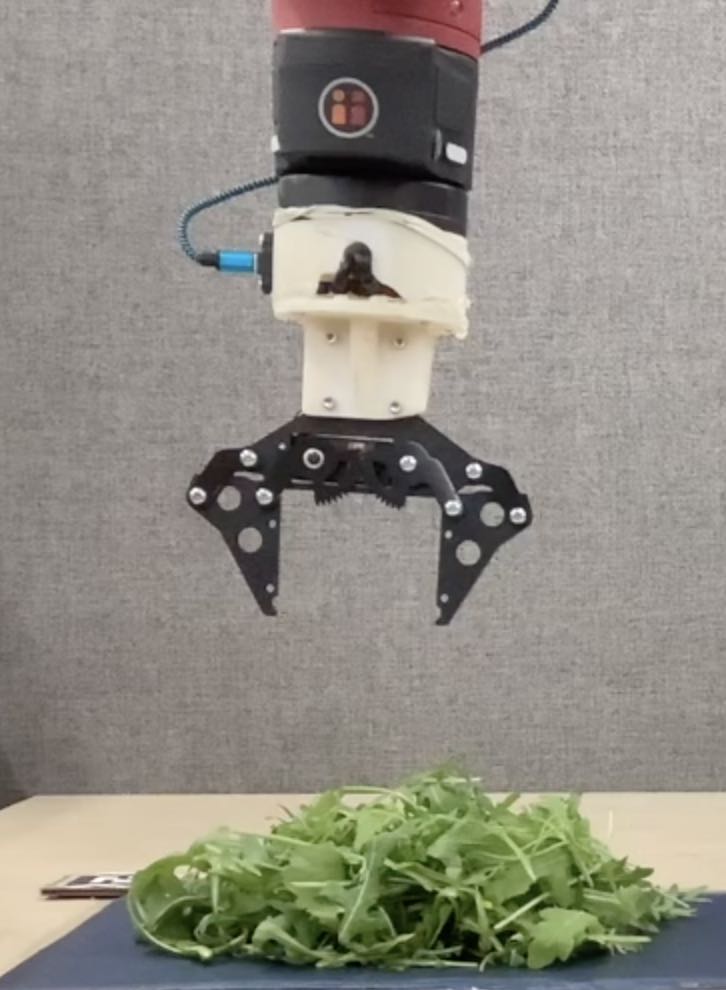}};%
\filldraw[color=orange] (-7.7,0.7) circle (4pt);%
\filldraw[color=red] (-8,1.3) circle (4pt);%
\draw [solid, white, line width=0.5mm] (-8.8,0.7) -- (-5.8,0.7);%
\draw [dashed, white, line width=0.5mm] (-7.1 ,-0.4) -- (-8.7,2.7);%
%
\draw [->,yellow, line width=0.5mm]  (-6.8,2.0) -- (-7.9,1.3);%
\note{-6.5,2}{Entanglement Point}%
%
\draw [->,yellow, line width=0.5mm] (-7.9,-0.3) -- (-7.8,0.6);%
\note{-7.8,-0.5}{Collision-free Point}%
\draw [solid, white, line width=0.5mm] (-4.4,0.7) -- (-1.4,0.7);%
\draw [dashed, white, line width=0.5mm] (-2.5 ,-0.4) -- (-4.3,2.7);%

\draw [->,line width=1.5pt] (-2.4,1.5)  arc [start angle=40,end angle=-235,x radius=0.95,y radius=0.24];

\draw [->, red, line width=0.5mm] (-1.35,2.5) -- (-1.35,3);%
\draw (-1.65,2.9) node {z};%
\draw [->, green, line width=0.5mm] (-1.35,2.5) -- (-1.85,2.5);%
\draw (-1.95,2.5) node {x};%
\node at (-9.27,-1.65){\colorbox{white}{\ref{f:top_plastic_herbs}}};%
\node at (-4.45,-1.65){\colorbox{white}{\ref{f:robot_cs}
}};%
\end{tikzpicture}%
\caption{Overview of the proposed \SP approach. %
\cl{%
 \item\label{f:top_plastic_herbs} Top view of the pile. %
\item\label{f:robot_cs} Front view showing the gripper. The solid white line represents the initial orientation of the $x$-axis of the gripper. The dashed white line represents the \emph{line of entanglement}. The black curved arrow represents the direction of rotation. Once the collision-free and entanglement points are identified, the gripper is rotated around the $z$-axis such that it aligns with the line of entanglement.  %
}%
}%
\label{f:overall_method_overview}%
\end{figure}%

\section{Method}\label{s:spread_and_pick}
Picking a target mass or number of tangle-prone \glspl{GM} is highly challenging due to the variability induced by the tangling. Although some level of tangling is unavoidable in the materials considered here, it is proposed to \emph{reduce this through a \SP strategy}. \fref{overall_method_overview} illustrates how the proposed approach works. In the first step, the location of a \emph{collision-free point} is estimated from an image of the grasping scene as a picking location. This helps to reduce the risk of damage to the plant material by minimising contact with the gripper, but usually still leads to variable picking mass due to tangling. Therefore, in the second step, the peak \emph{entanglement point} is estimated, and used to perform a spreading action such that the target mass is separated from the rest of the pile. The following describes how these points are estimated through a vision-based approach.


\subsection{Collision-free Gripper Pose: Graspability Index}\label{s:graspability_index}
The \gls{GI} \cite{Domae2014FastGrippers} is a vision-based measure for evaluating candidate grasping poses that has proved useful in industrial pick and place settings. It uses a single depth map of the scene to estimate the optimal gripper position and orientation for picking an object. It can be applied for use with different hand mechanisms, including parallel, multi-finger and vacuum grippers. It is particularly suitable for the picking problem considered here since it is unaffected by colour variation (that may occur between different plants) since only a depth map and a 2D gray-scale image are needed to process the scene. It should be noted, however, that its use of depth maps means it is most effective when a perpendicular view of the scene is available. 

For an insertion depth $\rz$, \GI estimates a point $\br$ %
in the bin such that the parallel plates of the gripper could be inserted without colliding with the objects inside. A range of $\rt$ is evaluated using GI and for the optimal $\rt$, the best picking point ($\rx$, $\ry$) is estimated. %

\begin{figure*}[t!]%
\centering
\begin{tikzpicture}%
\node [draw=none] at (0,0)%
{\includegraphics[width=.113\textheight]{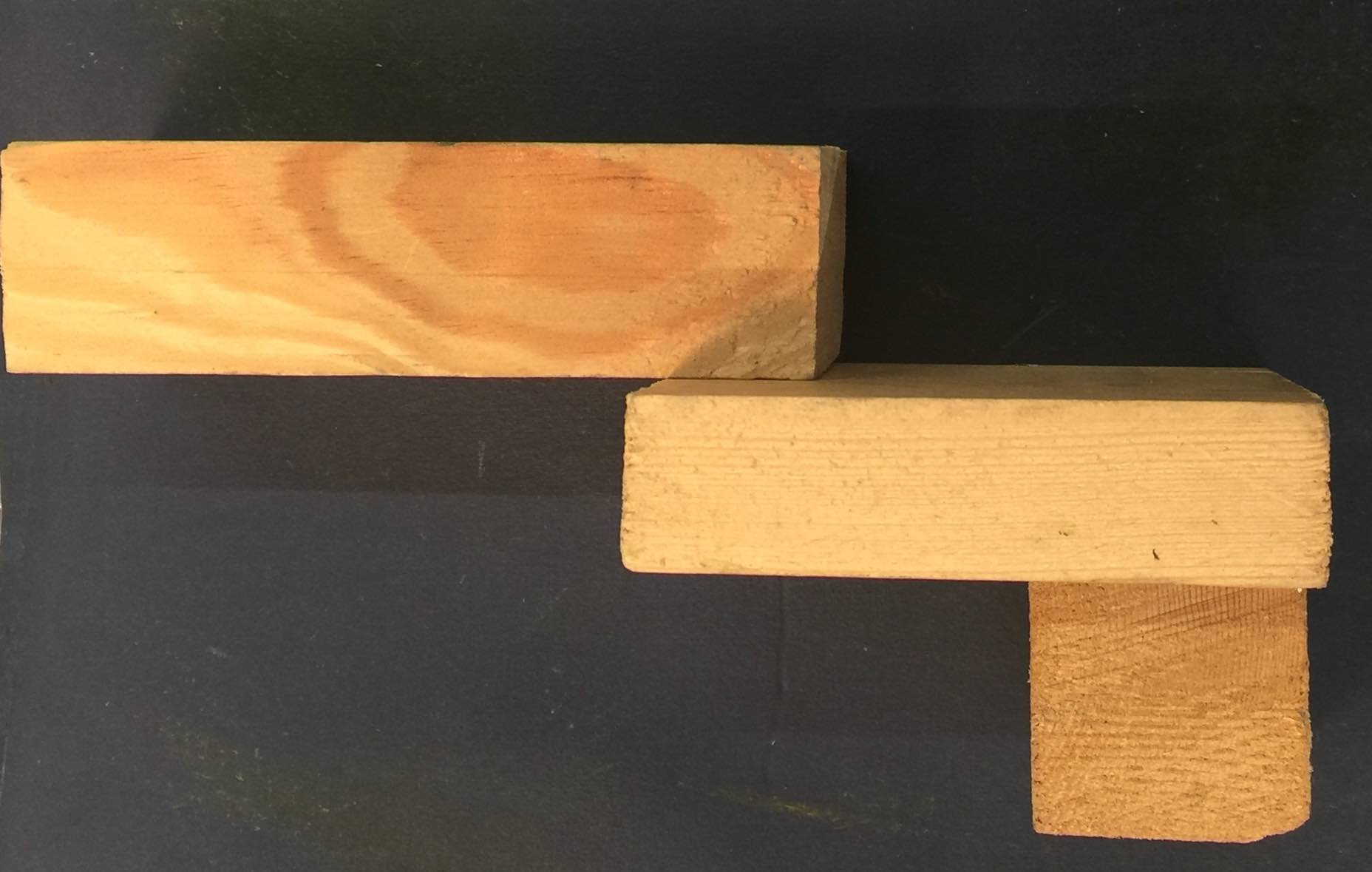} };%
\draw [->, yellow, line width=0.75mm] (-0.75,1.25) -- (1.00,0.050);%
\note{-0.75,1.25}{Target Object}%
%
\node [draw=none] at (2.8,2) {\includegraphics[width=.100\textheight]{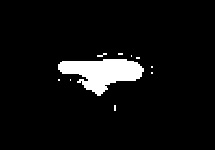} };%
\draw [-] (0,1.3) -- (0,2);%
\draw [->] (0,2) -- node [text width=3.25cm,left,above,align=left ] {\hspace{-.65cm}\parbox{\linewidth}{\footnotesize\raggedleft Binary image from depth map with \emph{highest point}\\as threshold.}} (0.9,2);%
%
\node [draw=none] at (2.8,-2) {\includegraphics[width=.100\textheight]{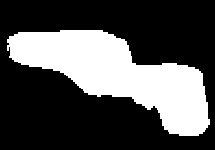} };%
\draw [-] (0,-1.3) -- (0,-2);%
\draw [->] (0,-2) -- node [text width=3.25cm,left,below,align=left ] {\hspace{-.65cm}\parbox{\linewidth}{\footnotesize\raggedleft Binary image from depth map with \emph{insertion depth} as threshold.}} (0.9,-2);%
%
%
%
\node at (4.5,2){$*$};%
%
\node [draw=none] at (5.8,2) {\includegraphics[width=.070\textheight]{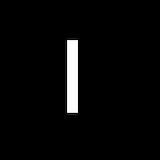} };%
\node at (5.8,0.7){$\gc$};%
%
%
\node at (4.5,-2){$*$};%
%
%
\node [draw=none] at (5.8,-2) {\includegraphics[width=.070\textheight]{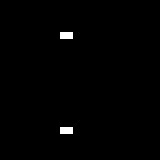} };%
%
%
\draw [->] (6.9,2) -- (7.3,2);%
\node [draw=none] at (8.8,2) {\includegraphics[width=.100\textheight]{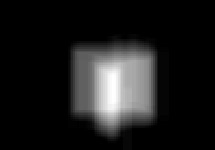}};%
%
%
\draw [->] (6.9,-2) -- (7.3,-2);%
\node [draw=none] at (8.8,-2) {\includegraphics[width=.100\textheight]{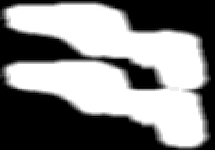}};%
%
%
%
\draw [->] (10.25,2) -- (10.65,2);%
\node at (13.8,2){$*$};%
\node [draw=none] at (12.1,2.0) {\includegraphics[width=.100\textheight]{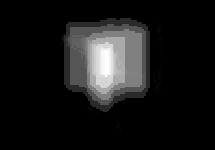}};%
\node at (12,1.35){$\wc \cap \overline{\wcp}$};%
%
\node [draw=none] at (14.8,2.2) {\includegraphics[width=.050\textheight]{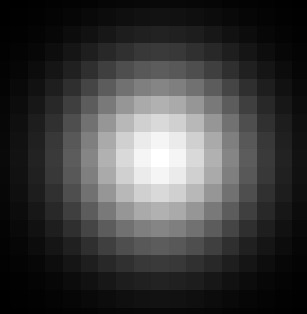}};%
%
%
\draw [->] (14.8,1.4) -- (14.8,1.15);%
\node [draw=none] at (14.9,-0) {\includegraphics[width=.110\textheight]{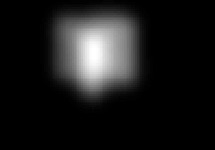}};%
%
%
\draw [->] (10.25,-2) -- (10.65,-2);%
\draw [->] (12.2,-0.2) -- (12.2,0.2);%
\node [draw=none] at (12.1,-2) {\includegraphics[width=.100\textheight]{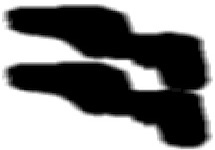}};%
%
%
\node at (-0.8,-1.3){\colorbox{white}{(\subref{f:gi_method_1}) \textbf{Scene}}};%
\node at (2.9,0.7){\colorbox{white}{(\subref{f:gi_method_2}) $\Oc$}};%
\node at (5.9,0.7){\colorbox{white}{(\subref{f:gi_method_4}) $\gc$}};%
\node at (9.0,0.7){\colorbox{white}{(\subref{f:gi_method_6}) $\wc$}};%
\node at (12.0,0.7){\colorbox{white}{(\subref{f:gi_method_8}) $\wc \bigcap \overline{\wcp}$}};%
\node at (15,-1.35){\colorbox{white}{(\subref{f:gi_method_10}) $\G$}};%
\node at (2.9,-0.7){\colorbox{white}{(\subref{f:gi_method_3}) $\Ocp$}};%
\node at (5.9,-0.7){\colorbox{white}{(\subref{f:gi_method_5}) $\gcp$}};%
\node at (9.0,-0.7){\colorbox{white}{(\subref{f:gi_method_wc}) $\wcp$}};%
\node at (12.05,-0.7){\colorbox{white}{(\subref{f:gi_method_7}) $\overline{\wcp}$}};%
\node at (16.0,2.20){\colorbox{white}{(\subref{f:gi_method_9}) $\g$}};%
{\phantomsubcaption\label{f:gi_method_1}}
{\phantomsubcaption\label{f:gi_method_2}}
{\phantomsubcaption\label{f:gi_method_3}}
{\phantomsubcaption\label{f:gi_method_4}}
{\phantomsubcaption\label{f:gi_method_5}}
{\phantomsubcaption\label{f:gi_method_6}}
{\phantomsubcaption\label{f:gi_method_wc}}
{\phantomsubcaption\label{f:gi_method_7}}
{\phantomsubcaption\label{f:gi_method_8}}
{\phantomsubcaption\label{f:gi_method_9}}
{\phantomsubcaption\label{f:gi_method_10}}
\end{tikzpicture}%
\caption{Estimating the grasping position ($\rx, \ry$) using \GI for gripper rotation $\rt = \ang{90}$. The scene contains three wooden blocks. In this example, the highest object (middle block) is the target object and the insertion depth $\rz$ is set such that the tips of the gripper just touch the surface of the table. The collision-free pick-up point $\Or$ is estimated from the peak of the graspability map $\G$.%
}%
\label{f:GI_Method}%
\end{figure*}
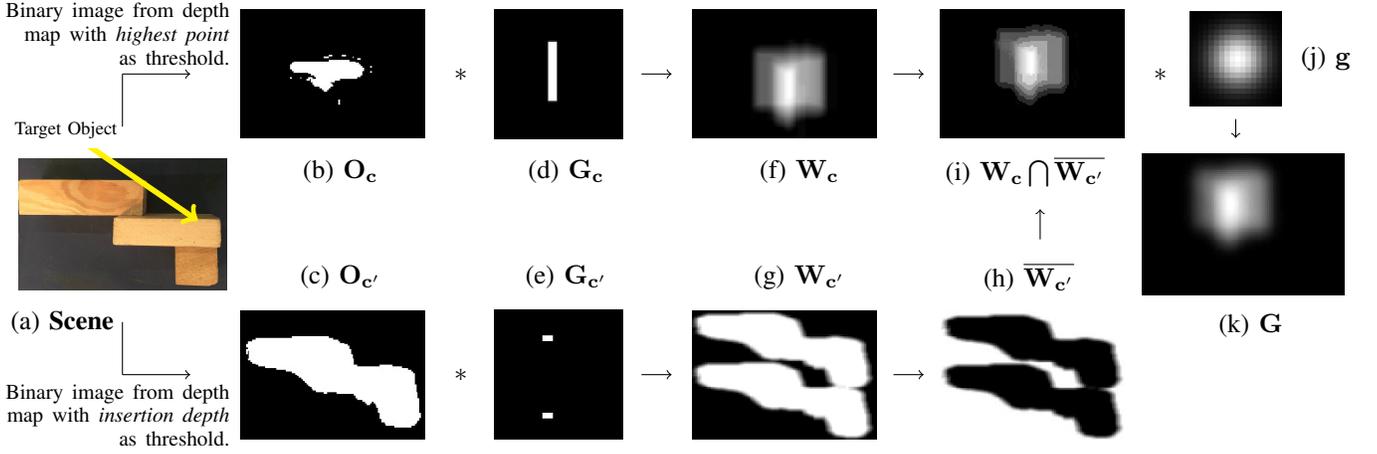%
	
\fref{GI_Method} provides an overview of the \GI method. First, a depth map of the cluttered scene is acquired using vision (\eg RGB-D camera). $\mathbf{\Oc}$ (see \fref{GI_Method}(\subref{f:gi_method_2})) represents the region of the target object that should lie between the gripper plates for a successful grasp. It is obtained by thresholding the depth map by the \emph{height of the target object} (middle block in \fref{GI_Method}(\subref{f:gi_method_1})). $\mathbf{\Ocp}$ represents the region in which a collision might occur while the gripper is moving downwards. It is obtained by thresholding the depth map by the \emph{insertion depth} $\rz$ (see \fref{GI_Method}(\subref{f:gi_method_3})). $\mathbf{\gc}$ and $\mathbf{\gcp}$ (see \fref{GI_Method}(\subref{f:gi_method_4}) and (\subref{f:gi_method_5}), respectively) represent the contact distance between the parallel plates and collision regions (\ie lateral width of the plates) for the gripper and are obtained through millimetre-to-pixel unit conversion. They are recomputed whenever the opening aperture of the gripper changes. The region where part of the target object lies between the gripper plates (\fref{GI_Method}(\subref{f:gi_method_6})) is computed through the convolution\footnote{Here, and throughout the paper, $\conv$ represents the convolution operation.}
\begin{equation}
\mathbf{\wc = \Oc \conv \gc}.
\label{e:gripper_object_contact}
\end{equation}
Similarly, the region where the gripper plates could collide with the objects in the pile is obtained as (see \fref{GI_Method}(\subref{f:gi_method_wc}))
\begin{equation}
	\mathbf{\wcp = \Ocp \conv \gcp.}
	\label{e:gripper_object_collision}
\end{equation}

The region of interest for successful picking is the area where contact between the gripper plates and the target object is detected and there is no collision with other objects in the bin.  Since $\mathbf{\wcp}$ represents the region where collisions might occur the latter may be expressed as ($\mathbf{\wc \cap \overline{\wcp}}$), where the notation $\overline{\mathbf{A}}$ represents the \emph{NOT} operation on $\mathbf{A}$ and $\cap$ denotes intersection (see \fref{GI_Method}(\subref{f:gi_method_8})). Finally, using a Gaussian $\mathbf{\g}$ (see \fref{GI_Method}(\subref{f:gi_method_9})), the graspability map  $\mathbf{\G}$ is computed as 
\begin{equation}
	\mathbf{G =  (\wc \cap \overline{\wcp})  \conv g}.
	\label{e:gripper_object_collision-1}
\end{equation}
Convolution with a Gaussian $\g$ is used to smooth and reduce the noise in the graspability map. The peak of $\mathbf{\G}$ is obtained for a range of gripper orientations $\rt$ to determine the respective  pick up point ($\rx$, $\ry$) by maximising
\begin{equation}
	f(x,y,\rt)=\begin{cases}
		\mathbf{(\G})_{xy}, & \text{if } \mathbf{(\wcp})_{xy} {=} 0\\
		0, & \text{otherwise}.
	\end{cases}
	\label{e:graspability_index_target_function}
\end{equation}
where $\mathbf{(\G})_{xy}$ and $\mathbf{(\wcp})_{xy}$ represents the value of $\G$ and $\wcp$ at position $(x,y)$ respectively. Gripper orientations for which no peak could be detected are discarded and $\rt$ is set to the the gripper orientation for which the peak could be determined in $\mathbf{\G}$ yielding the picking position
\begin{equation}
	\Or =(\rx,\ry,\rt)^\T
	 = \operatorname*{argmax}_{x, y} f(x, y,\rt).
	\label{e:graspability_index_4}
\end{equation}

The optimal gripper position and orientation as obtained from the \GI identify a reference for the gripper for collision-free picking of the target object. However, this
ignores the possibility that parts of the target object could be entangled with other items in the bin such that it may end up picking them along with the target. In case of herbs, experience tells that this frequently occurs resulting in more than the desired mass being picked. In the next section, a strategy is proposed for \emph{reducing tangling during the pick operation} to help alleviate this problem.

\subsection{Tangle Reduction}\label{s:collision_region}
To reduce the level of tangling and thereby achieve more consistent picking, this paper proposes a \SP approach, inspired by human behaviour. In humans, it is frequently observed that they use their fingers to separate things while picking, especially when they have to work with one hand. The idea here is to mimic this behaviour by adjusting the pick to include a spreading step: specifically, if the target object is between the plates of the gripper, instead of moving them inwards (closing) to grasp the object, they are first moved \emph{outward} to try to disentangle any nearby objects before proceeding with the pick.

The proposed approach extends the \GI by identifying regions of high entanglement in the scene and then defining a spreading movement to disentangle them. For a specific $\rt$, $\mathbf{\gcp}$ is used to obtain $\mathbf{\wcp}$, the region that represents gripper-object collision. $\mathbf{\wcp}$ is then used to identify the region of entanglement

\begin{equation}
	\mathbf{\G' = \wcp \conv g .}
	\label{e:extended_graspability_index}
\end{equation}

Using $\mathbf{\G'}$, the \emph{peak entanglement position} is computed as

\begin{equation}
	\Er = ({\rx},{\ry},{\rt})^\T
	 = \operatorname*{argmax}_{x, y} h(x, y, \rt)
	\label{e:graspability_index_3}
\end{equation}
where 
\begin{equation}
	h(x,y,\rt)=\begin{cases}
		(\G')_{xy}, & \text{if } (\wcp)_{x y}{=}1\\
		0, & \text{otherwise}.
	\end{cases}
	\label{e:extended_graspability_index_target_function}
\end{equation}

The \emph{line of peak entanglement} is then defined as that intersecting $\Er$ and $\Or$. This line defines the spreading movement in the proposed approach: during the pick operation, the gripper plates are moved outwards along this line to disperse the tangle and improve the consistency of picking. \PL, being an informative feature in the design of the picking strategies, can also be used to adjust the proposed outward movement considering the tangling propensity of different \glspl{GM}. \fref{Extended_GI_Method_timelapse} illustrates the working of the robot while following the \SP approach.

\begin{figure*}[t!]
\centering%
\begin{tikzpicture}
\node [draw=none] at (0,0) {\includegraphics[width=.170\textheight]{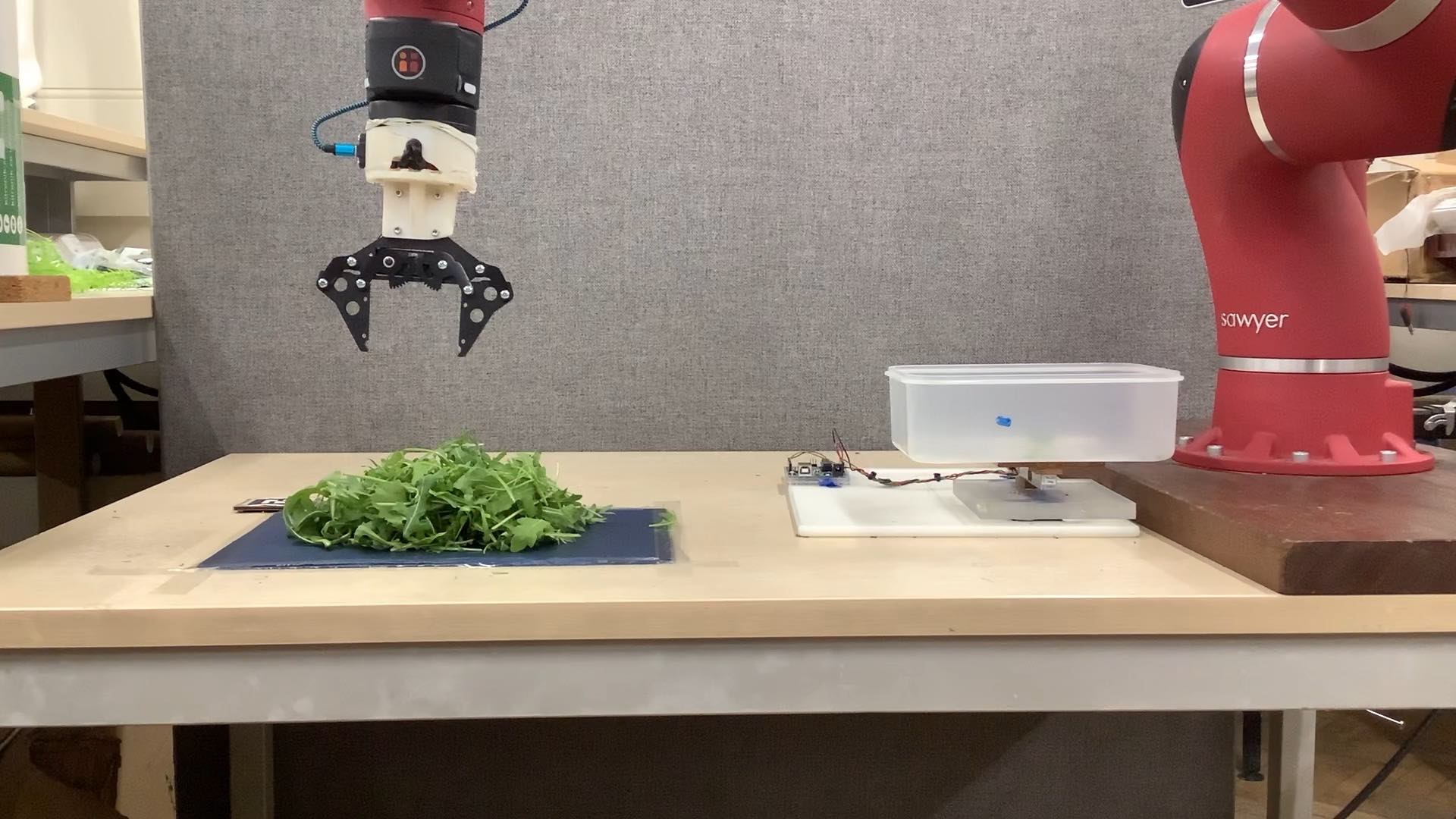} };
\node [draw=none] at (4.5,0) {\includegraphics[width=.170\textheight]{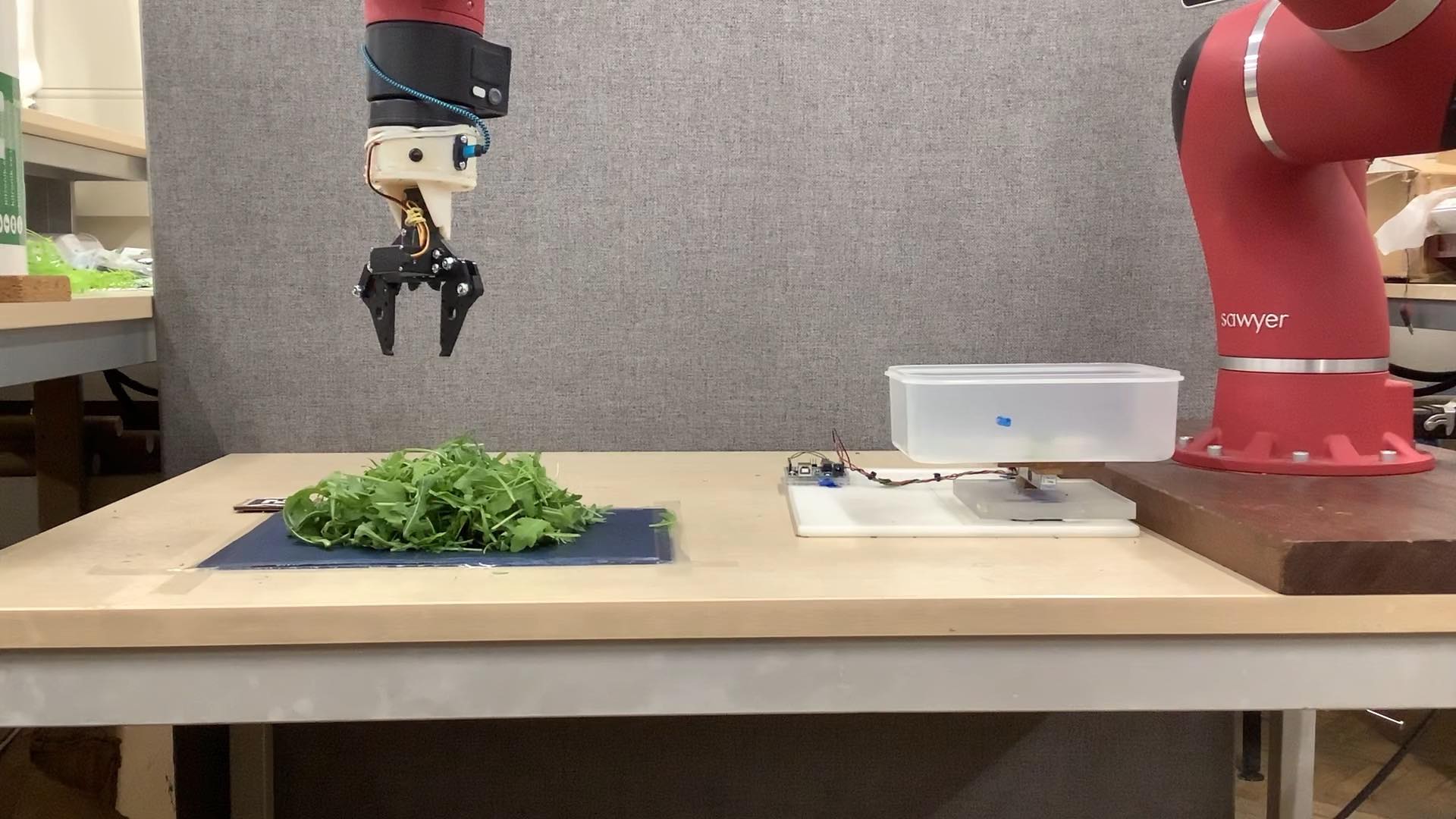} };
\node [draw=none] at (9,0) {\includegraphics[width=.170\textheight]{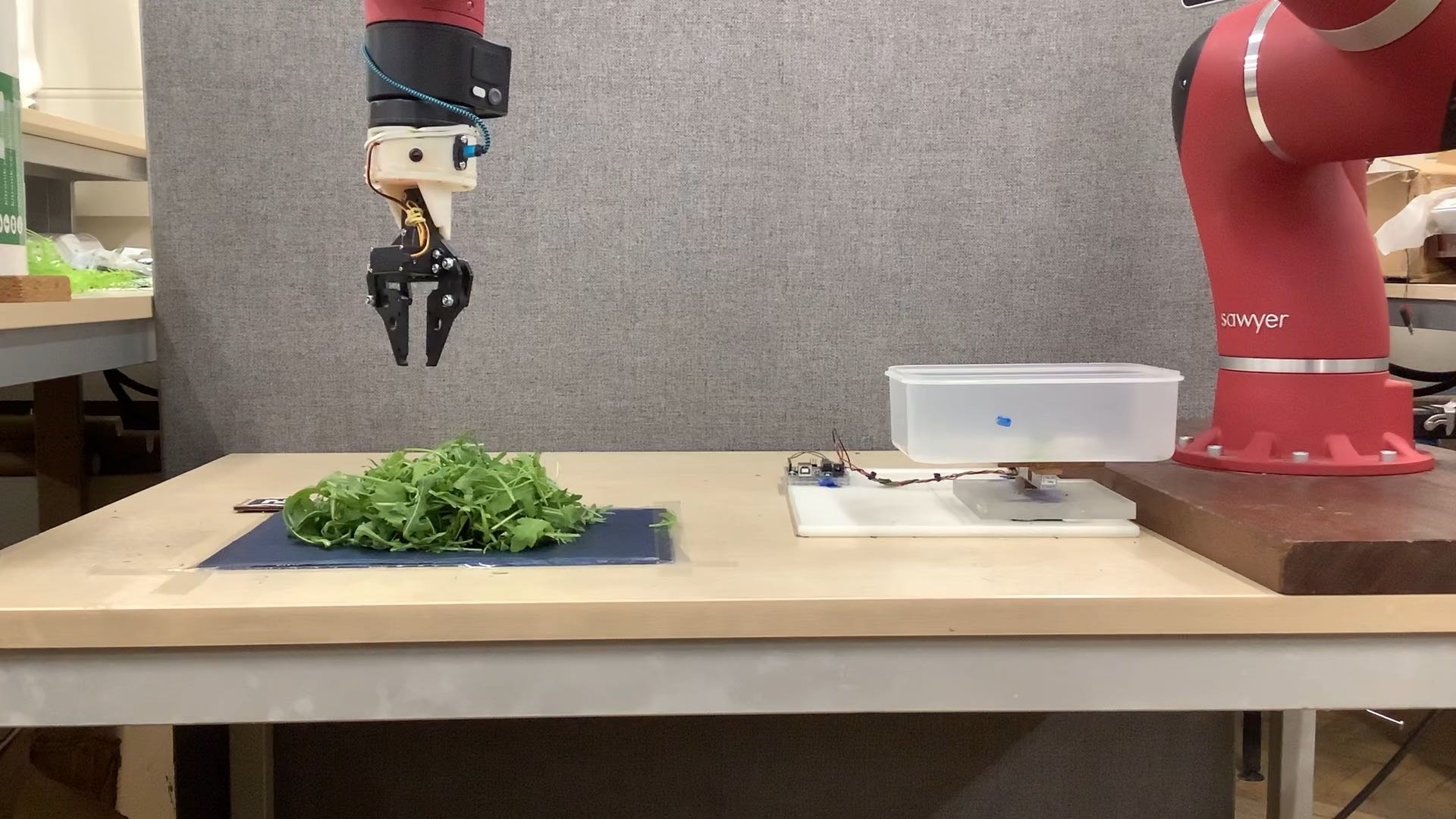} };
\node [draw=none] at (13.5,0) {\includegraphics[width=.170\textheight]{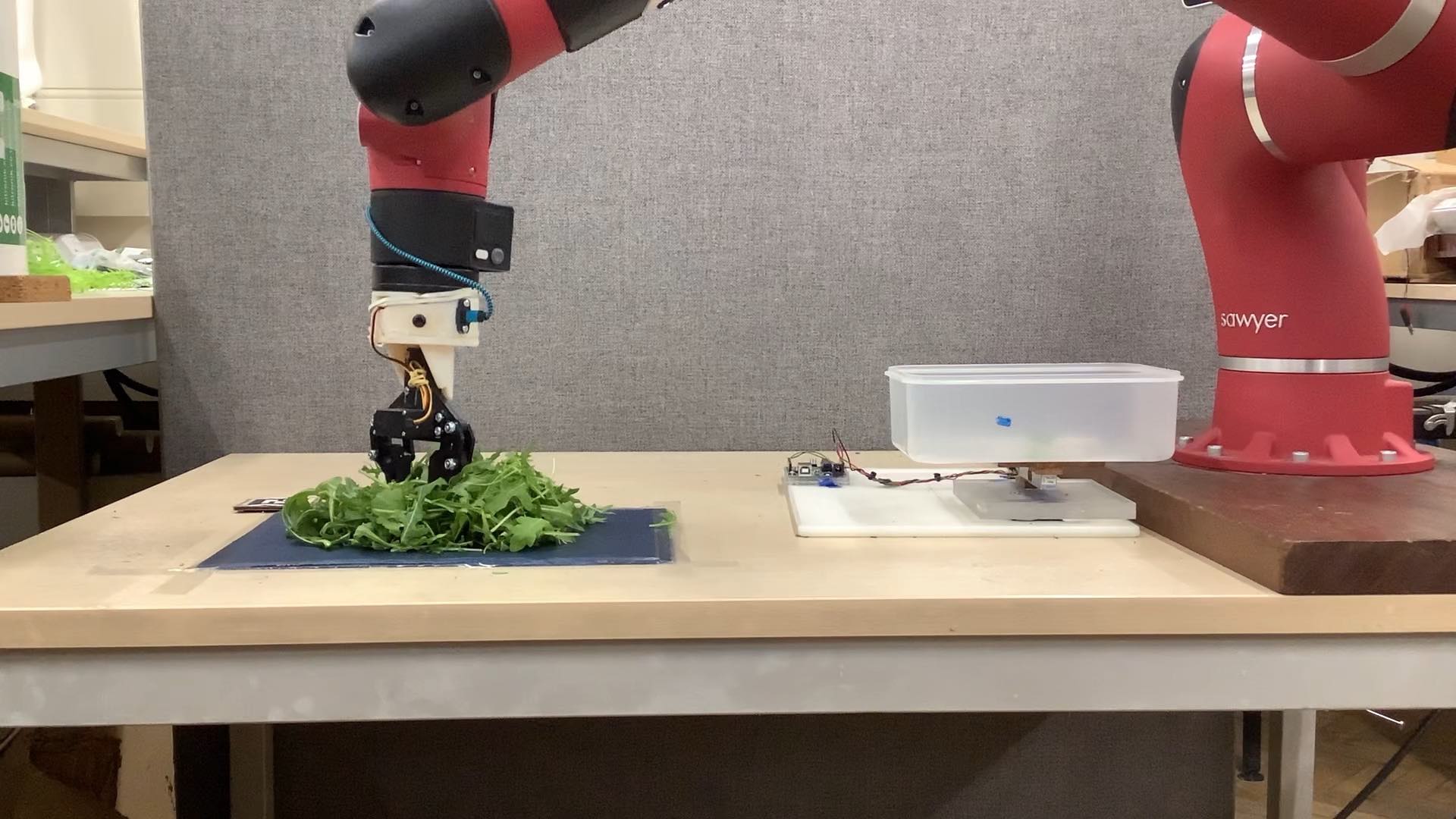} };

\node [draw=none] at (0,-2.5) {\includegraphics[width=.170\textheight]{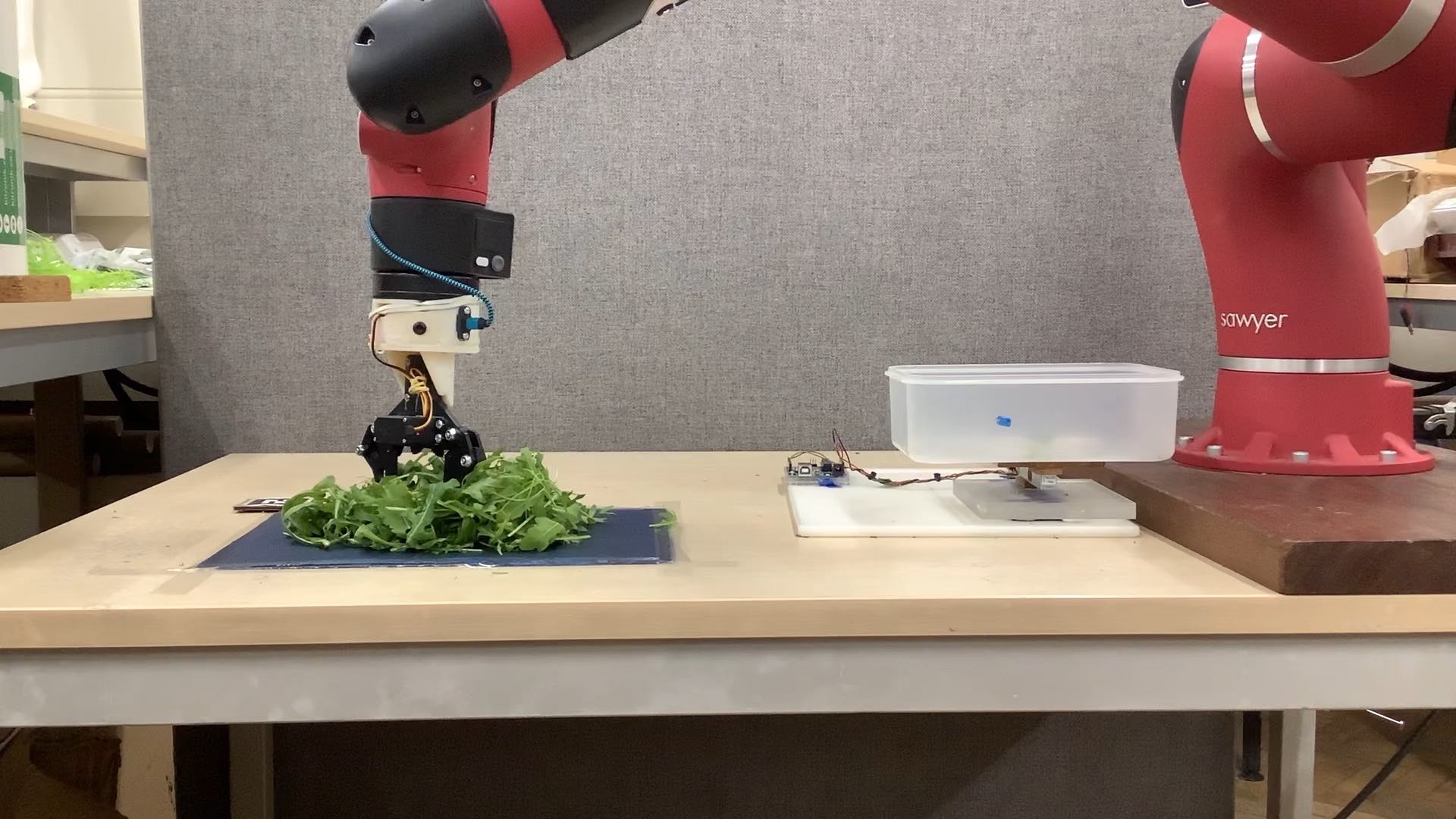} };
\node [draw=none] at (4.5,-2.5) {\includegraphics[width=.170\textheight]{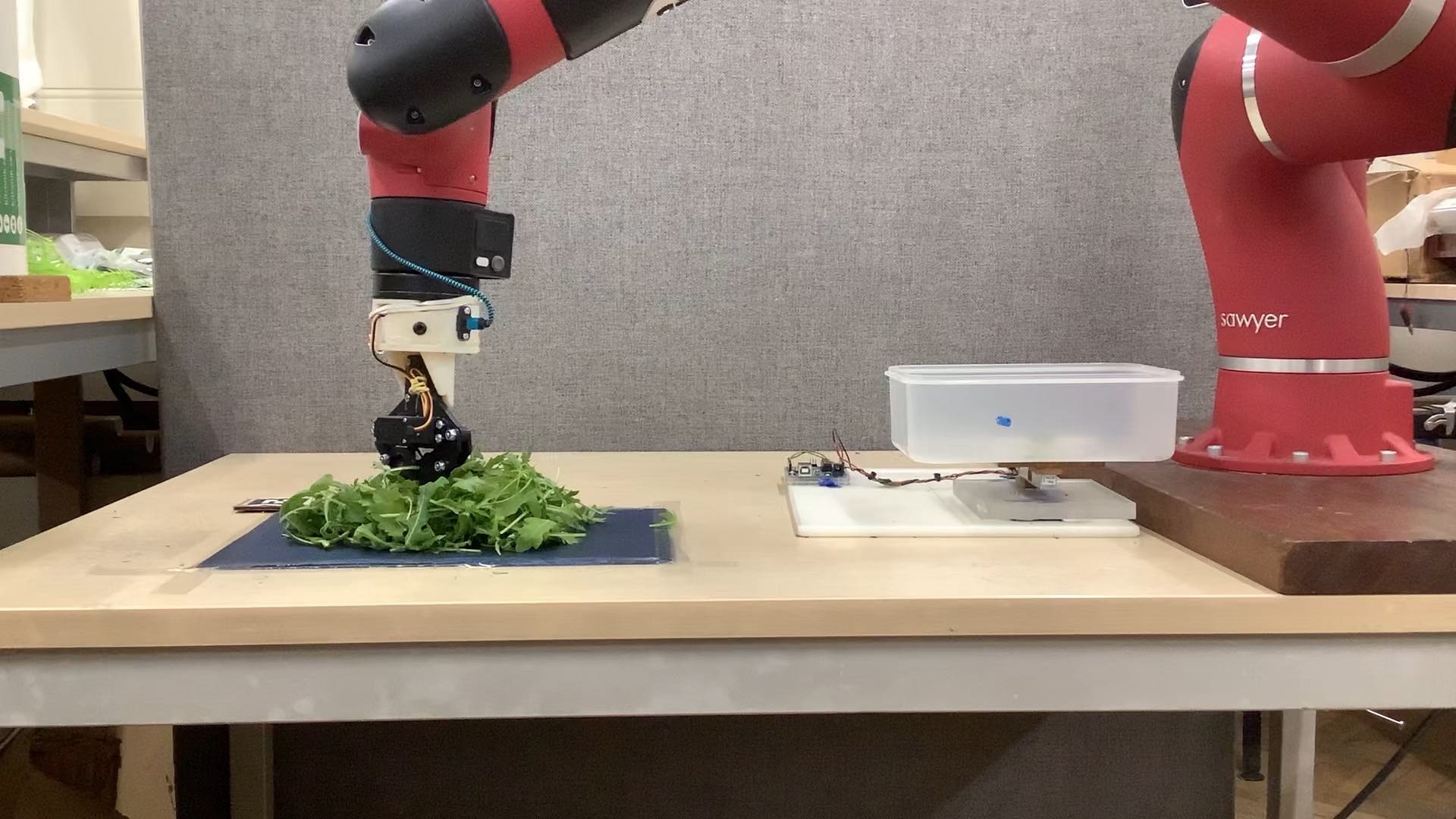} };
\node [draw=none] at (9,-2.5) {\includegraphics[width=.170\textheight]{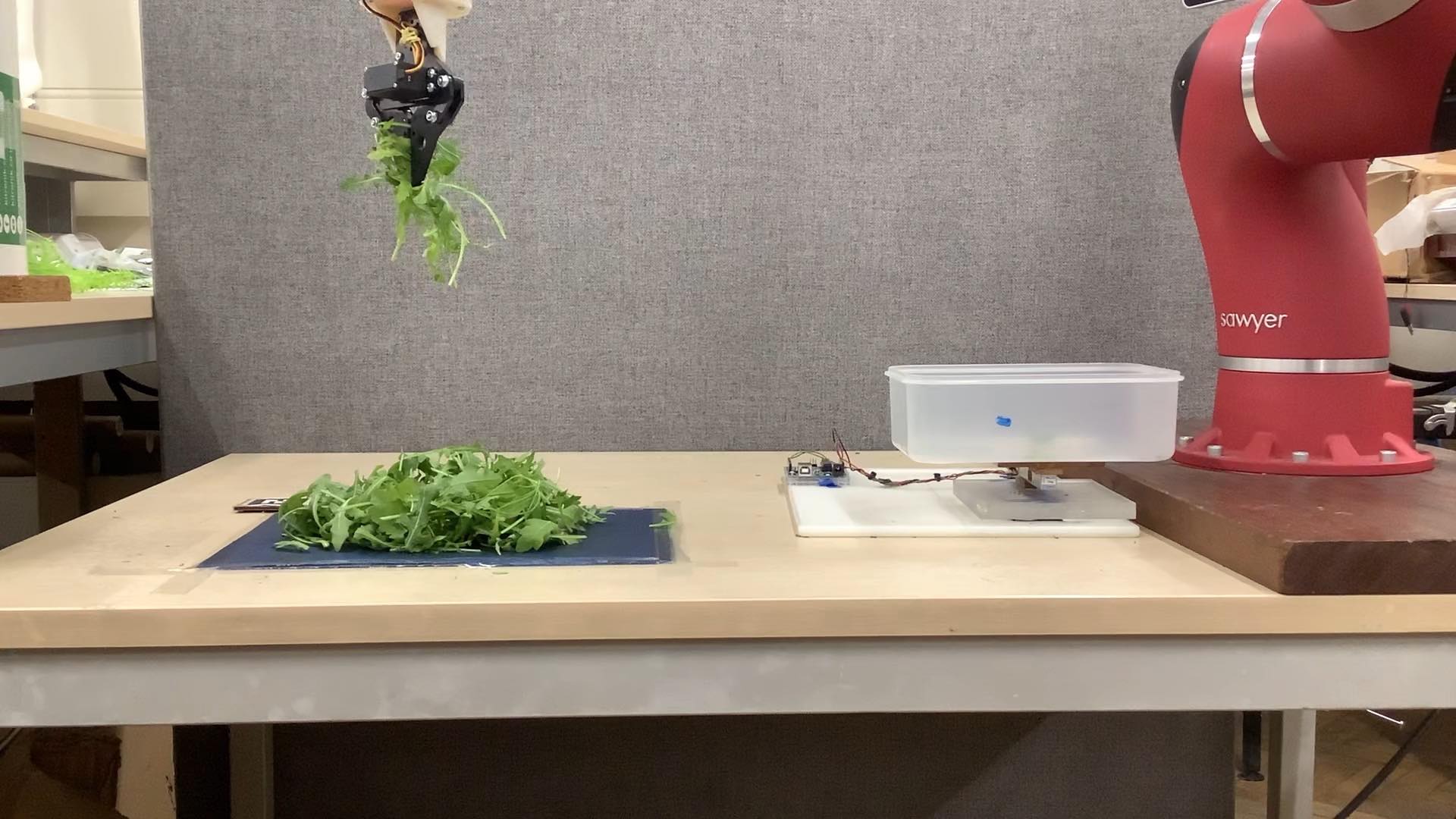} };
\node [draw=none] at (13.5,-2.5) {\includegraphics[width=.170\textheight]{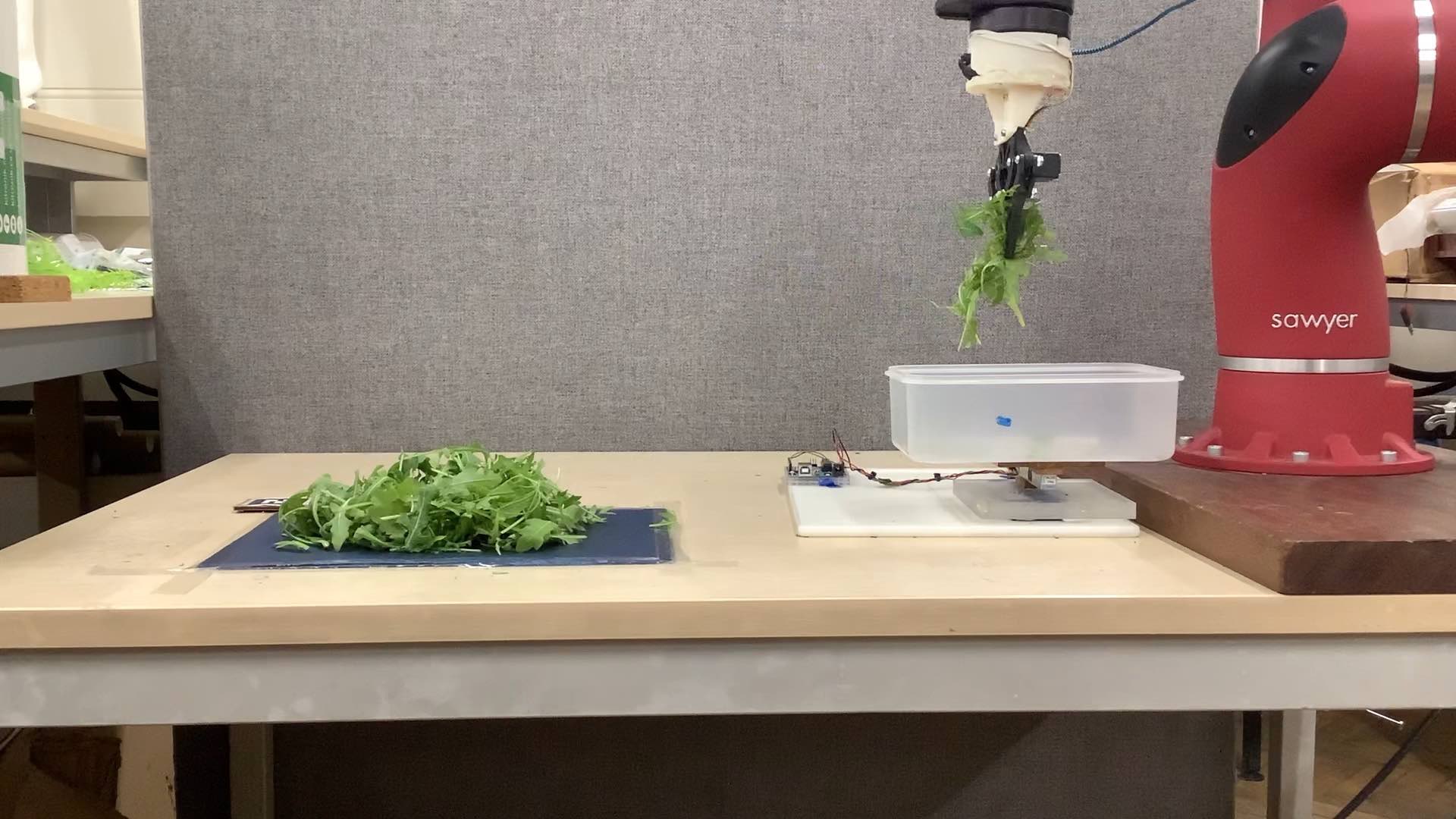} };

\node at (-1.79,-0.92){\colorbox{white}{\ref{f:extended_method_overview_1}}};
\node at (2.72,-0.92){\colorbox{white}{\ref{f:extended_method_overview_2}}};
\node at (7.2,-0.92){\colorbox{white}{\ref{f:extended_method_overview_3}}};
\node at (11.7,-0.92){\colorbox{white}{\ref{f:extended_method_overview_4}}};

\node at (-1.79,-3.43){\colorbox{white}{\ref{f:extended_method_overview_5}}};
\node at (2.70,-3.43){\colorbox{white}{\ref{f:extended_method_overview_6}}};
\node at (7.2,-3.43){\colorbox{white}{\ref{f:extended_method_overview_7}}};
\node at (11.7,-3.43){\colorbox{white}{\ref{f:extended_method_overview_8}}};
\end{tikzpicture}

\caption{Time lapse illustrating \SP approach. 
\cl{%
\item\label{f:extended_method_overview_1} Robot reaches a fixed point above the pile.%
\item\label{f:extended_method_overview_2} Gripper orientation adjusted to align with \emph{line of peak entanglement.} %
\item\label{f:extended_method_overview_3} Gripper aperture set to chosen width.%
\item\label{f:extended_method_overview_4} Gripper moved into herb pile to pick from the optimal collision-free point according to \GI. %
\item\label{f:extended_method_overview_5} Gripper plates moved outwards to maximum aperture width.%
\item\label{f:extended_method_overview_6} Gripper closed.%
\item\label{f:extended_method_overview_7} Gripper raised with items picked.%
\item\label{f:extended_method_overview_8} Picked items dropped onto scale to record mass. %
}%
}%
\label{f:Extended_GI_Method_timelapse}
\end{figure*}
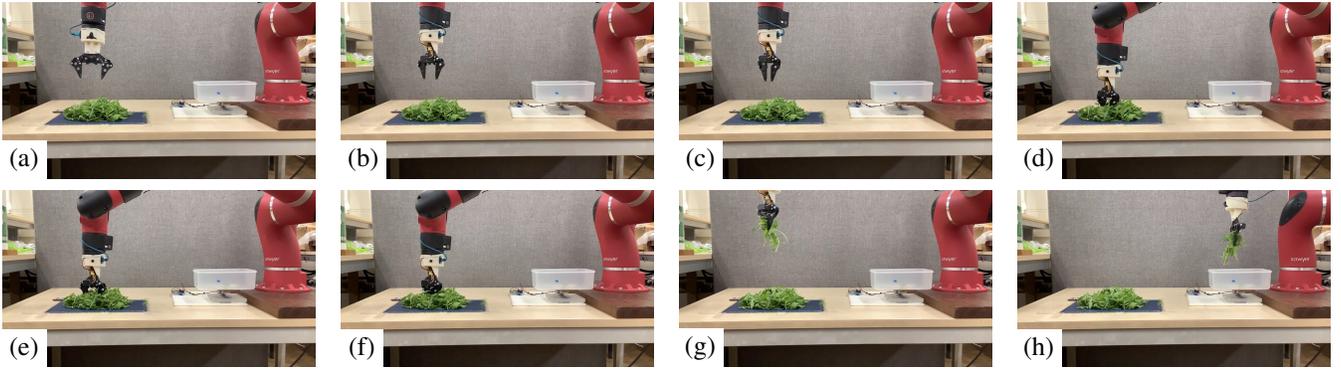


\subsection{Mass-constrained Picking}\label{s:mass-constrained-picking}
The purpose of \SP is to reduce the pile entanglement to a reasonable threshold such that the remaining element of $\pickpar$ (\ie gripper aperture $\w$) can be estimated efficiently for any target mass $\mt$. The training set $\mathbf{\mathcal{D}}$ to learn the skill as expressed in \eref{pickpar_eq} is collected by running $\mathcal{N}$ picking trials for different pick parameter $\pickpar$ and consists of a matrix of pick parameters $(\pickpar_1,....,\pickpar_p)\in\R^{4\times \mathcal{P}}$
and the corresponding matrix of observed picked masses $\big(\begin{smallmatrix}
  m_{1}^{\pickpar_1} &...& m_{1}^{\pickpar_p}\\
  .&.&.\\
  m_n^{\pickpar_1}&...&m_n^{\pickpar_p}\\
\end{smallmatrix}\big) \;\in\;\mathbb{R^{\mathcal{N}\times \mathcal{P}}}$.

This data $\mathbf{\mathcal{D}}$ is used to fit a predictive model through supervised learning. The inverse of the model is then used to achieve the desired skill \eref{pickpar_eq} as presented in \sref{problem_def}. It should be noted, however, that the chosen model should be monotonic for to be inverted.

\section{Experiments}
In this section, three experimental studies\footnote{The data supporting this research are openly available from
the King’s College London research data repository, KORDS, at \url{https://doi.org/10.18742/19977779}. Further information about the data and conditions of
access can be found by emailing \url{research.data@kcl.ac.uk}} are presented. The first validates the role that protrusions play in causing entanglement in two example tangle-prone \glspl{GM}. The second two experiments evaluate the proposed \gls{SnP} method for improving picking consistency for these materials.

The experimental set up is a mock-up of the packing workstation of a large fresh herbs and salads producer equipped with a robotic manipulator (see \fref{Experimental_Setup}). As the robotic platform, a $7$-\DoF Rethink Robotics Sawyer is used, with a maximum reach of $\pm\SI{1260}{\milli\meter}$ and precision of $ \SI{0.1}{\milli\meter}$. For simplicity and lower cycle-time,  $3$-\DoF of the robot are used for picking movements. The robot is equipped with a parallel gripper from Actobotics (product code: 637092) as its end-effector. 
The latter has maximum opening aperture of $w=\SI{71.12}{\milli\meter}$ and is controlled using a Hitec HS-422 Servo Motor with operating voltage range  $\SI{4.8}{\volt}$-$\SI{6.0}{\volt}$. As the vision module, the platform uses an Intel realsense d435i depth camera  mounted on a stand at a fixed position and orientation with respect to the robot. For simplicity of image processing, the camera position is chosen such that its field of view exactly covers the picking area and it records depth data at a frequency of $\SI{15}{\hertz}$. The mass picked is recorded using a parallel beam type load cell with a combined error of $\pm{0.05}\%$ and maximum weighing capacity of $\SI{10}{\kilogram}$. A HX711 amplifier combined with an Arduino microcontroller is used for data acquisition from the load cell.

The two tangle-prone \glspl{GM} considered here are \il{\item staples (a homogeneous \gls{GM}) and \item cut herbs and baby-leaf salads (a non-homogeneous \glspl{GM})}. %
For the former, general purpose office staples manufactured by Rapesco Office Products (923 type staples) are used. These are manufactured in different sizes and thus provide the flexibility of varying the protrusion lengths (\PLs) in a controlled manner, while ensuring homogeneity (each staple in the pile is identical). For the latter, either plastic or real herbs and salads are used. Use of plastic herbs enables some degree of control against natural variations in the real herbs, or changes in their physical properties (\eg due to plant material drying out, or becoming damaged over successive picks). Real herbs are used in the final evaluation to assess usability in a practical industrial application.

\subsection{Do protrusions cause tangling?}
\label{s:fp_plastic_herbs}
The first experiment tests the hypothesis:
\begin{enumerate*}[label=$\mathbf{H_\arabic*}$]
	\item\label{h1}\emph{The presence of protrusions leads to entanglement.}
\end{enumerate*}
For this, a simple picking task is designed to compare the picking variance for \glspl{GM} with and without protrusions. Two varieties of plastic herbs are chosen as the \gls{GM} for this experiment: one with no protrusions and the other having many protrusions of varied lengths extending from a central stem (see \fref{plastic_with_without_ends} \ref{f:no_transverse_ends} and \ref{f:transverse_ends}, respectively). The experimental procedure is as follows.

\subsubsection{Procedure}\label{s:fp_plastic_herbs_procedure}
Using the experimental set up illustrated in \fref{Experimental_Setup}, a series of robotic picking operations are conducted. During the experiment, a fixed mass of herbs are placed in a pile in an open picking area of dimension $\SI{30}{\centi\meter}\times\SI{25}{\centi\meter}$. Each picking operation consists of the robot reaching into the pile as per the pick parameter $\pickpar$, closing its gripper, and lifting what is grasped free of the surface.  In detail, in each pick, the gripper orientation is initialised to $\rt=\ang{90}$, target picking location ($\rx$, $\ry$) is fixed as the center of the pile and the insertion depth $\rz$ is set such that the tips of the gripper just touch the surface of the picking area. The robot moves its end-effector to a fixed position above the picking area,  sets the gripper aperture $w$ to the chosen width and lowers it into the pile. %
There, it closes the gripper plates, moves its end-effector vertically upwards to a fixed position, and drops what has been picked into the weighing device to record the mass. To ensure a similar physical arrangement of the plant material between trials, any material picked is returned, the entire quantity is manually transferred to a $\SI{18}{\centi\meter}\times\SI{13.5}{\centi\meter}\times\SI{7}{\centi\meter}$ cuboid container and then replaced onto the picking area for the next pick. For each type of plant material used, picking is conducted $30$ times for gripper aperture $\w \in \{20, 30, 40, 50, 60\}\SI{}{\milli\meter}$ and pile mass $\p={30}\SI{}{\gram}$.


\subsubsection{Results}\label{s:results_h1}
\fref{fp_plastic_result} reports the picked mass as observed for the \gls{GM} with and without protrusions. As can be seen, average picked mass for the \gls{GM} with protrusions is higher for all $\w$  as compared to \gls{GM} without protrusions, confirming \ref{h1} as higher degree of tangling causes more to be picked than is expected. Additionally, the variance for the \gls{GM} with protrusions is considerably higher, than for those without for all $\w$ %
with a maximum increase of 76\% in picked mass variance for \gls{GM} with protrusions.

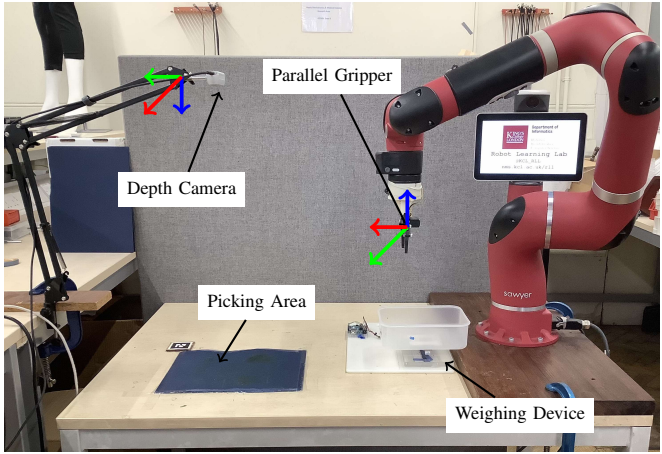
\begin{figure}[t!]%
	\centering%
	\begin{annotate}{\includegraphics[width=0.48\textwidth]{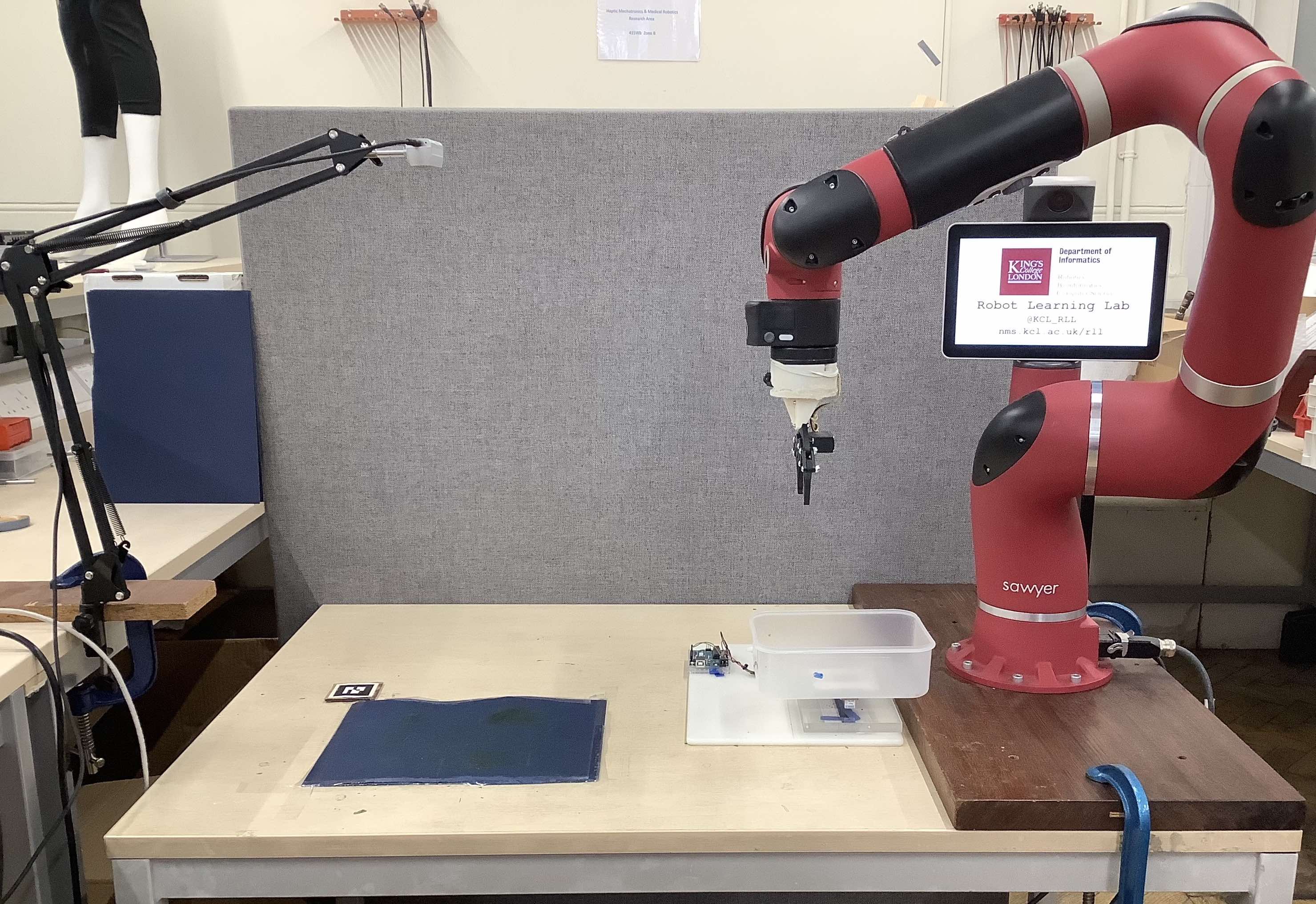}}{1}%
		\callout{-2,0.5}{Depth Camera}{-1.5,1.8}%
		\callout{-1,-1}{Picking Area}{-1.5,-1.8}%
		\callout{2.5,-2.5}{Weighing Device}{1.5,-1.8}%
		\callout{0,2}{Parallel Gripper}{1,0}%
		\draw [->, blue, line width=0.5mm] (-2,2) -- (-2,1.5);%
		\draw [->, green, line width=0.5mm] (-2,2) -- (-2.5,2);%
		\draw [->, red, line width=0.5mm] (-2,2) -- (-2.5,1.5);%
		\draw [->, blue, line width=0.5mm] (1,0) -- (1,0.5);%
		\draw [->, green, line width=0.5mm] (1,0) -- (0.5,-0.5);%
		\draw [->, red, line width=0.5mm] (1,0) -- (0.5,0);%
	\end{annotate}%
	\caption{Overview of the experimental set up. Red, green and blue arrows represent $x$-, $y$- and $z$-axes, respectively. The coordinate frame attached to the robot is used as the frame of reference.}%
	\label{f:Experimental_Setup}%
\end{figure}%

\subsection{Does \gls{SnP} improve picking consistency?}
\label{s:sp_staples}
To evaluate the effectiveness of the \SP approach as compared to \FP, the next experiment tests the hypothesis: 
\setcounter{enumi}{1}%
\begin{enumerate*}[label=$\mathbf{H_\arabic*}$,resume]
\item\label{h2}\emph{Picking following \SP results in a significant increase in picking consistency as compared to \FP-based picking}.
\end{enumerate*}

In this experiment, more precise control of the factors with a possible effect on tangling is required, so staples with constant staple width $d=\SI{12}{\milli\meter}$ and variable protrusion length $l$ (see \fref{plastic_with_without_ends} \ref{f:one_staple}) are chosen as the \gls{GM} for this experiment: each staple is identical with \il{\item only two protrusions, \item fixed protrusion length and \item fixed volume, shape and density}. The experimental procedure is as follows.
\subsubsection{Procedure}
A similar procedure to that outlined in \sref{fp_plastic_herbs_procedure} is followed with one key difference: instead of an open area, picking is performed directly from a cuboid container of dimension $\SI{12.8}{\centi\meter}\times\SI{10.6}{\centi\meter}\times\SI{2}{\centi\meter}$  mounted on the weighing device, and the \GM is vibrated for $\SI{10}{\second}$ using a micro vibrator with rated voltage $\SI{3}{\volt}$ and rotating speed $\SI{12000} {RPM}$ prior to each pick. This eliminates the manual transfer of material in and out of the container between picks and helps ensure consistent packing of the material across picks. For \SP, after lowering its end-effector into the pile,  the robot performs the spreading manoeuvere, closes the gripper plates,  moves its end-effector vertically upwards to a fixed position,  records the picked mass and then drops what has been picked back into the container. The procedure is repeated $60$ times for sets of staples with \PLs $\l \in \{6, 8, 10,12\}\SI{}{\milli\meter}$, gripper aperture $\w = 40 \SI{}{\milli\meter}$ and pile mass $\p=\SI{60}{\gram}$. The same procedure is then repeated for \FP picking. Note that, in the latter, \il{\item no spreading movement is performed and \item the target picking location is fixed as the centre of the pile}.

\subsubsection{Results}\label{s:results_h2}
Table \ref{t:sp_gi_fp_results} reports the picked mass for the \FP and \SP methods. It is observed that the standard deviation of the picked mass is the least for \SP among all cases. This demonstrates that \SP reduces entanglement in the pile without having to precisely define and measure \emph{degree of entanglement}, making picking more consistent and hence confirming \ref{h2}.

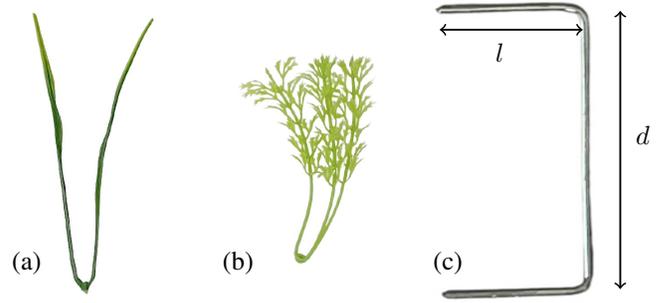
\begin{figure}[t!]
	\centering%
	\begin{tikzpicture}%
		\node  at (-11.5,0.60){\includegraphics[width=.12\textwidth]{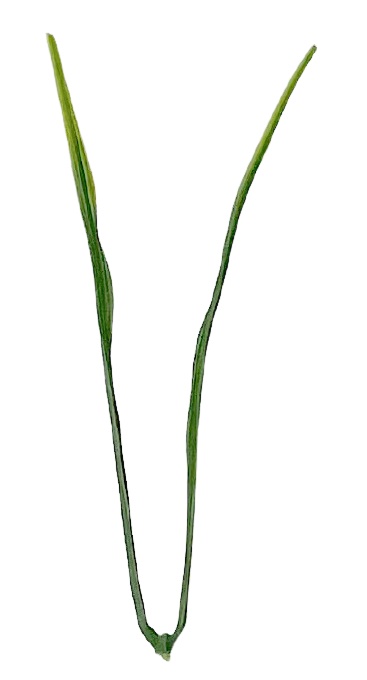}};
		\node  at (-8.7,0.60){\includegraphics[width=.12\textwidth]{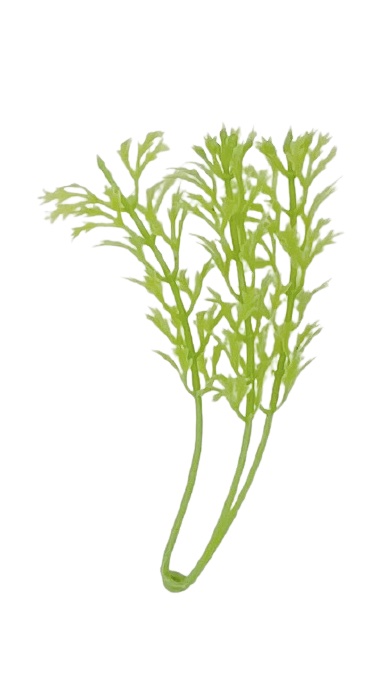}};
		\node  at (-5.9,0.60){\includegraphics[width=.12\textwidth]{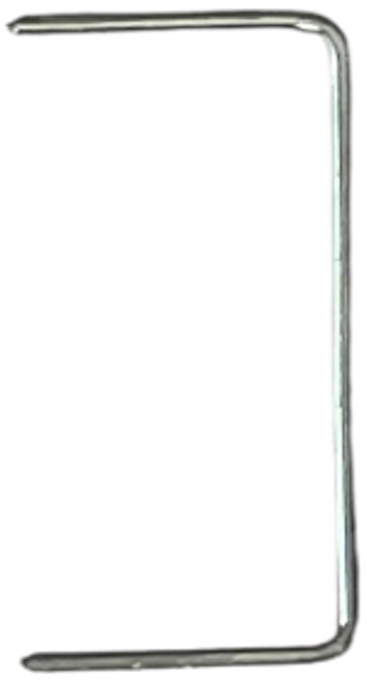}};
		
		\draw [<->, black, line width=0.25mm] (-4.5,-1.25) -- (-4.5,2.45);
		\node at (-4.2,0.8){$d$};
		\draw [<->, black, line width=0.25mm] (-6.9,2.2) -- (-5,2.2);
		\node at (-6.1,1.9){$l$};
		
		\node at (-12.39,-0.9){\colorbox{white}{\ref{f:no_transverse_ends}}};
		\node at (-9.58,-0.9){\colorbox{white}{\ref{f:transverse_ends}}};
		\node at (-6.79,-0.9){\colorbox{white}{\ref{f:one_staple}}};
		
	\end{tikzpicture}
	\caption{\glspl{GM} used in the experiments.
		\cl{%
			\item\label{f:no_transverse_ends} Plastic herbs without and %
			\item\label{f:transverse_ends} with protrusions,  and %
			\item\label{f:one_staple} Staples with constant staple width $d=\SI{12}{\milli\meter}$ and variable protrusion length $l$. %
		}%
	}%
	\label{f:plastic_with_without_ends}
\end{figure}

\subsection{Industrial herb and salad picking task}\label{s:industrial_herb_and_salad_picking_task}
In this section, the proposed \SP method is evaluated with respect to its efficacy in improving consistency in an industrial picking task, namely, picking a target mass of fresh herbs and salads (wild rocket and parsley, see \fref{herbs}). Specifically, the next experiment test the hypothesis:
\setcounter{enumi}{1}%
\begin{enumerate*}[label=$\mathbf{H_\arabic*}$,resume]
\item\label{h3}\emph{Picking following \SP results in a greater picking accuracy as compared to \GI-based picking}.
\end{enumerate*}


\subsubsection{Procedure}
Data is collected using the mock picking station rig shown in \fref{Experimental_Setup} and used to fit a predictive model of the required $\pickpar$ given a target masses for each \GM considered. Specifically, picking is performed $20$ times for gripper aperture $\w\in\{20, 30, 40, 50, 60\}\SI{}{\milli\meter}$ for plastic herbs and $10$ times for $w \in \{20, 30, 40\}\SI{}{\milli\meter}$ for real herbs, using the procedure outlined in \sref{fp_plastic_herbs_procedure}. This data is used to estimate a linear model, that is inverted to derive the skill as presented in \eref{pickpar_eq} for  
computing required gripper aperture for target mass $\mt$.
The remaining elements of $\pickpar$ (\ie picking location and orientation $\br$) are computed according to the procedure described in \sref{graspability_index} and \sref{collision_region} respectively. To evaluate the accuracy and consistency of picking, this method is applied to pick a series of target masses: $\mt\in\{8, 10, 12\}$\SI{}{\gram} for plastic herbs ($\mt\in\{15, 20\}$\SI{}{\gram} for real herbs) and the error (\ie absolute difference from actual mass picked) is recorded. This is repeated for $20$ trials ($10$ trials for real herbs). For comparison, the experiment is also repeated using standard \GI-based picking (\ie picking at the collision-free point, and omitting the spreading movement). To further test the robustness, the experiment is also repeated with the variation that the picking model for wild rocket is applied to picking material from a different plant, namely, flat-leaf parsley.

\subsubsection{Results}\label{s:case_study}
Tables \ref{t:normal_plastic_test_results} and \ref{t:normal_real_test_results} report the \e\ for picking plastic herbs and wild rocket, respectively. It is observed that the \e\ with \SP is lower among all cases, and is up to $41$\% lower for picking $\mt=\SI{10}{\gram}$ of plastic herbs and up to $19$\% for picking $\mt=\SI{20}{\gram}$ of wild rocket. A significant decrease in the \e\ variance is also observed with \SP for all cases. Finally, \tref{parsley_results} provides the \e\ for picking flat-leaf parsley using the model derived for wild rocket. As observed, the \e\ is lower for \SP compared to the \GI-based approach for all target masses considered with up to $51$\% for picking $\mt=\SI{20}{\gram}$. These results show that the proposed \SP approach effectively reduces picking error and improves picking consistency for a variety of herbs and salads.

\begin{figure}[t!]
\centering %
\begin{tikzpicture}%
\node at (0,0){\includegraphics[height=0.25\textheight]{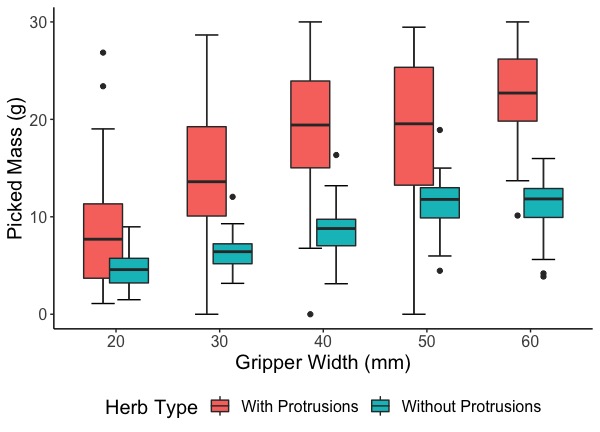}};
\end{tikzpicture}
\caption{Results from fixed-point picking (\FP) experiments involving plastic herbs with pile mass $p=\SI{30}{\gram}$, reporting \emph{picked mass} across $30$ trials.
}%
\label{f:fp_plastic_result}
\end{figure}

\section{Discussion}
This work studies \glspl{GM} in the context of robotic bin-picking,  a research focus that until now has been largely unexplored. The experimental results show strong evidence that \il{\item protrusions play an important role in causing mechanical entanglement in \GMs and \item that the proposed \SP approach is effective in reducing picking error and improving picking consistency for a variety of herbs and salads}. 

The results in \sref{fp_plastic_herbs} demonstrate an increase in picked mass variance for \GMs with protrusions, demonstrating their importance in causing tangling, and consequently decreased consistency in picking. This suggests that protrusion length \PL is an informative feature for achieving better generalisation, especially in a mass-constrained robot picking tasks involving a variety of \glspl{GM} with protrusions. 

In case of no entanglement (see \fref{SDandTransverseEnds}(\subref{f:full_ends_1jpeg.jpg}), (\subref{f:mid_ends_1jpeg.jpg}) and (\subref{f:no_ends_1jpeg.jpg})), only the target object is expected to be picked despite the decreasing protrusion length $\l$. In case of entanglement (see \fref{SDandTransverseEnds}(\subref{f:full_ends_4jpeg.jpg}), (\subref{f:mid_ends_3jpeg.jpg}) and (\subref{f:no_ends_11jpeg.jpg})), contact surface area decreases as $\l$ decreases and undesired objects are more likely to fall off---suggesting a monotonic relationship between degree of entanglement and $l$. However, it is interesting to note that for both \FP and \SP-based picking, the maximum standard deviation in picked mass is observed for staples with an intermediate protrusion length $\l=\SI{10}{\milli\meter}$ (see \tref{sp_gi_fp_results}). This suggests a non-monotonic relationship between the degree of entanglement and $l$. As reported in \cite{Gravish2012PhysicsRev}, this behaviour is attributed to the interplay between the \emph{ability to pack} and the \emph{ability to entangle} in a \gls{GM} pile.


\begin{table}[t!]%
    \tiny
	\captionof{table}{picked mass s.d. of staples (over 60 trials) for protrusion length $\l \in \{6, 8, 10,12\}\SI{}{\milli\meter}$, staple width $d=12 \SI{}{\milli\meter}$, gripper aperture $\w = 40 \SI{}{\milli\meter}$ and pile mass $\p=\SI{60}{\gram}$. \label{t:sp_gi_fp_results}}%
	\centering%
	\begin{adjustbox}{width=0.469\textwidth}%
		\begin{tabular}{ccc}%
			\hline
			$\l$ (\si{\milli\meter}) & \FP (\si{\gram})& \SP (\si{\gram}) \\ \hline
			$6$ & $1.013$ & $0.753$ \\ \hline
			$8$ & $1.370$ & $1.141$ \\ \hline
			$10$ & $1.729$ & $1.640$ \\ \hline
			$12$ & $1.333$ & $1.156$ \\ \hline
		\end{tabular}
	\end{adjustbox}
\end{table}

\begin{table}[t!]%
    \tiny
	\captionof{table}{\e\ in picking plastic herbs (mean$\pm$s.d. over 20 trials). \label{t:normal_plastic_test_results}}
	\centering%
	\begin{adjustbox}{width=\columnwidth}%
		\begin{tabular}{ccc}
			\hline
			$\mt$ (\si{\gram}) & Method & \e\ (\si{\gram}) \\ \hline
			\multicolumn{1}{c}{\multirow{2}{*}{$8$}} & \GI & $5.191\pm2.709$ \\
			\multicolumn{1}{c}{} & \SP & $3.772\pm2.655$ \\ \hline
			\multicolumn{1}{c}{\multirow{2}{*}{$10$}} & \GI & $6.516\pm6.408$   \\
			\multicolumn{1}{c}{} & \SP & $3.820\pm2.253$ \\ \hline
			\multicolumn{1}{c}{\multirow{2}{*}{$12$}} & \GI & 
			$6.995\pm5.959$ \\
			\multicolumn{1}{c}{} & \SP & $5.090\pm3.149$ \\ \hline
		\end{tabular}
	\end{adjustbox}
\end{table}



Moreover, the results from the study in \sref{sp_staples} and the industrial herb and salad picking task (\sref{industrial_herb_and_salad_picking_task}) demonstrate the power of the proposed \SP approach in reducing error and improving picking consistency by tackling tangling in \GMs. In the former, controlled staple-picking experiment, \SP results in a lower standard deviation of the picked mass in all cases, demonstrating the important effect that reducing tangling can have. Moreover, in the herb/salad picking task the \e\ is shown to be lower when using \SP in all cases. Interestingly, comparing the \e\ for plastic and real herbs, the reduction in \e\ is lower for the former. This difference is attributed to factors such as moisture variation and a generally higher degree of entanglement in the real herbs. It is worth noting that,   the real herb material occasionally prevented the gripper plates from fully opening due to their tendency to tangle around the gripper itself. The presence of moisture in real plant material also tends to cause adhesion between herb strands in addition to the mechanical entanglement, potentially exacerbating the effect. Surprisingly, the maximum decrease in \e\ is observed for target mass $\mt=20$\SI{}{\gram} for flat-leaf parsley (see \tref{parsley_results}), even though gripper aperture was estimated using the wild rocket picking model. The interplay between the \emph{ability to pack} and the \emph{ability to entangle} is considered responsible for such an observation.

Overall,  it can be seen that the proposed \SP method proves effective in directly countering tangling in a variety of \GMs. It is not practical to train a separate model for all individual \glspl{GM} and the insights regarding protrusions as presented in this work, provide an useful way of generalising a picking model, especially considering the physical properties of the \glspl{GM}.  
\begin{figure}[t!]
\centering%
\begin{tikzpicture}

\draw [->, black, line width=0.75mm] (-10,1.7) -- (-2,1.7);
\node at (-6,2.1){\emph{decreasing protrusion length $\l$}};

\node [draw] at (-9,0) {\includegraphics[width=.100\textheight]{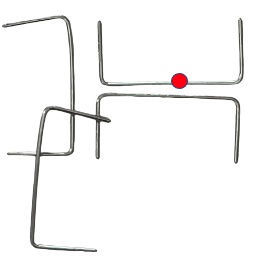} };
\node [draw] at (-6,0) {\includegraphics[width=.100\textheight]{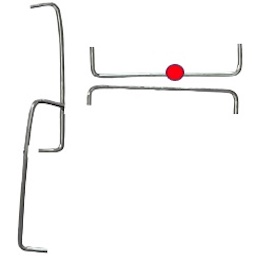} };
\node [draw] at (-3,0) {\includegraphics[width=.100\textheight]{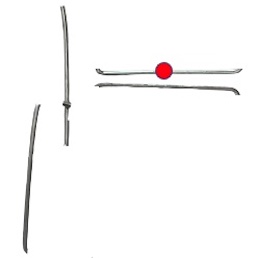} };

\node [draw] at (-9,-3) {\includegraphics[width=.100\textheight]{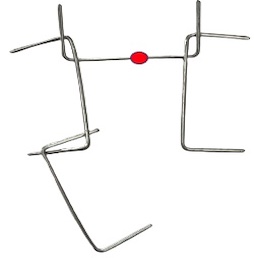} };
\node [draw] at (-6,-3) {\includegraphics[width=.100\textheight]{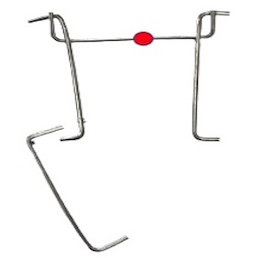} };
\node [draw] at (-3,-3) {\includegraphics[width=.100\textheight]{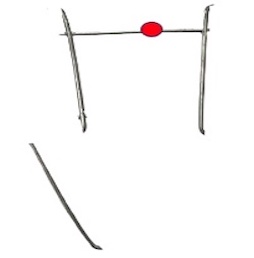} };

\node at (-8.1,-1.0){\colorbox{white}{(\subref{f:full_ends_1jpeg.jpg})}};
\node at (-5.1,-1.0){\colorbox{white}{(\subref{f:mid_ends_1jpeg.jpg})}};
\node at (-2.1,-1.0){\colorbox{white}{(\subref{f:no_ends_1jpeg.jpg})}};

\node at (-8.1,-4.0){\colorbox{white}{(\subref{f:full_ends_4jpeg.jpg})}};
\node at (-5.1,-4.0){\colorbox{white}{(\subref{f:mid_ends_3jpeg.jpg})}};
\node at (-2.1,-4.0){\colorbox{white}{(\subref{f:no_ends_11jpeg.jpg})}};

{\phantomsubcaption\label{f:full_ends_1jpeg.jpg}}%
{\phantomsubcaption\label{f:mid_ends_1jpeg.jpg}}%
{\phantomsubcaption\label{f:no_ends_1jpeg.jpg}}%
{\phantomsubcaption\label{f:full_ends_4jpeg.jpg}}%
{\phantomsubcaption\label{f:mid_ends_3jpeg.jpg}}%
{\phantomsubcaption\label{f:no_ends_11jpeg.jpg}}%

\end{tikzpicture}

\caption{Without ((\subref{f:full_ends_1jpeg.jpg}), (\subref{f:mid_ends_1jpeg.jpg}) and (\subref{f:no_ends_1jpeg.jpg})) and with entanglement ((\subref{f:full_ends_4jpeg.jpg}), (\subref{f:mid_ends_3jpeg.jpg}) and (\subref{f:no_ends_11jpeg.jpg})) scenarios. The red dot represents the target object.}

\label{f:SDandTransverseEnds}
\end{figure}

\section{Future Work}
This work considers piles composed of \glspl{GM} with varied protrusion lengths \PLs. Incorporating \PLs in the separation strategy can further improve the usefulness of \SP in reducing the pile entanglement.  Studying other tangle-prone \glspl{GM} such as rings would prove valuable for developing a more general framework for reducing entanglement in a pile. Future work might consider exploring other representations for the desired picking skill. A representation of the degree of entanglement in terms of the \emph{ability to pack} and the \emph{ability to entangle} could also be studied in the future for developing a generalised model suitable for a variety of challenging \glspl{GM}.  Additionally, this work acknowledges that entanglement is often unavoidable for \glspl{GM} causing more mass than desired to be picked up despite use of \SP. In such cases, it may help to consider in-hand strategies to remove unwanted materials such as by learning how to drop the extra undesired material without repeating the pick and using non-standard hand mechanisms.  

\begin{table}[t!]
\tiny%
	\captionof{table}{\e\ in picking wild rocket (mean$\pm$s.d. over 10 trials).\label{t:normal_real_test_results}}
	\centering%
	\begin{adjustbox}{width=\columnwidth}%
		\begin{tabular}{ccc}
			\hline
			$\mt$ (\si{\gram}) & Method & \e\ (g) \\ \hline
			\multicolumn{1}{c}{\multirow{2}{*}{$15$}} & \GI & $6.008\pm3.133$ \\
			\multicolumn{1}{c}{} & \SP & $5.091\pm2.533$ \\ \hline
			\multicolumn{1}{c}{\multirow{2}{*}{$20$}} & \GI & $6.529\pm5.495$ \\
			\multicolumn{1}{c}{} & \SP & $5.297\pm3.414$ \\ \hline
		\end{tabular}
	\end{adjustbox}
\end{table}

\begin{table}[t!]
\tiny%
	\captionof{table}{\e\ in picking flat-leaf parsley (mean$\pm$s.d. over 10 trials). Gripper aperture $\w$ are estimated using the wild rocket model. \label{t:parsley_results}}
	\centering%
	\begin{adjustbox}{width=\columnwidth}%
		\begin{tabular}{ccc}
			\hline
			$\mt$ (g) & Method & \e\ (g) \\ \hline
			\multicolumn{1}{c}{\multirow{2}{*}{$15$}} & \GI &  $7.444\pm4.365$ \\
			\multicolumn{1}{c}{} & \SP & $6.464\pm3.466$\\ \hline
			\multicolumn{1}{c}{\multirow{2}{*}{$20$}} & \GI & $7.535\pm5.863$ \\
			\multicolumn{1}{c}{} & \SP & $3.729\pm2.257$ \\ \hline
		\end{tabular}
	\end{adjustbox}
\end{table}


%




\section*{Acknowledgment}
The authors would like to thank Vitacress Herbs Ltd. for funding a part of this study. 

\begin{figure}[t!]
\centering%
\begin{tikzpicture}%
\node at (-11.5,0.60){\includegraphics[width=.12\textwidth]{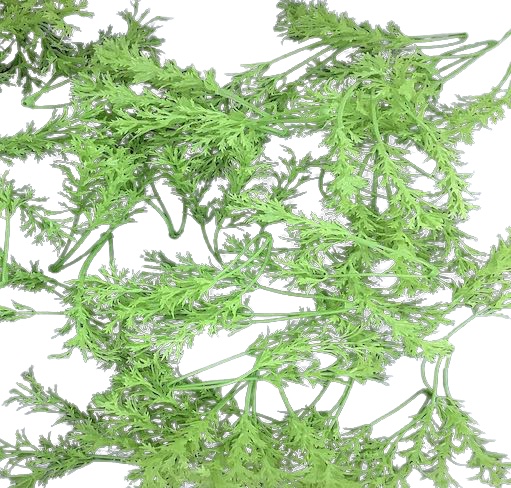}};
\node  at (-8.7,0.60){\includegraphics[width=.12\textwidth]{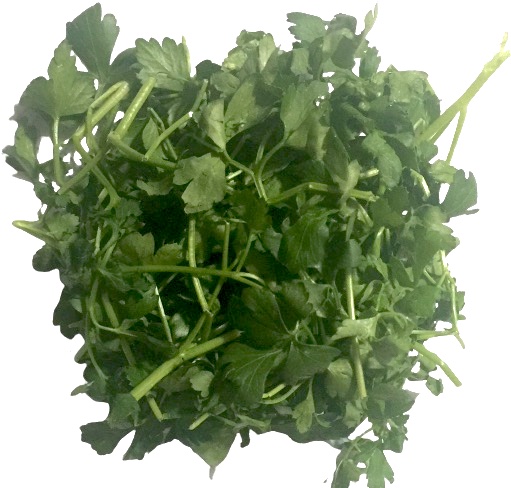}};
\node  at (-5.9,0.60){\includegraphics[width=.12\textwidth]{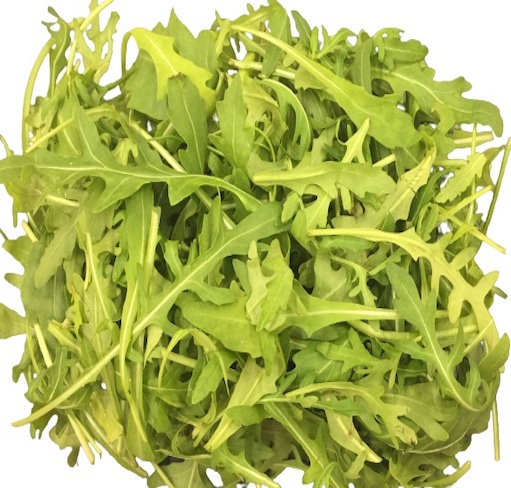}};

\node at (-12.39,-0.28){\colorbox{white}{\ref{f:bunch_plastic_herbs}}};
\node at (-9.58,-0.28){\colorbox{white}{\ref{f:bunch_parsley}}};
\node at (-6.79,-0.28){\colorbox{white}{\ref{f:bunch_wild_rocket}}};

\end{tikzpicture}
	\caption{Example \GMs used in experiments: \cl{%
			\item\label{f:bunch_plastic_herbs} plastic herbs %
			\item\label{f:bunch_parsley} parsley and %
			\item\label{f:bunch_wild_rocket} wild rocket. %
			}%
		}%
	\label{f:herbs}
\end{figure}
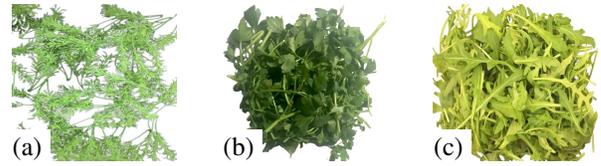


\ifCLASSOPTIONcaptionsoff
  \newpage
\fi



%

\scriptsize{\printbibliography}%

%


\begin{IEEEbiography}[{\includegraphics[width=1in,height=1.25in,clip,keepaspectratio]{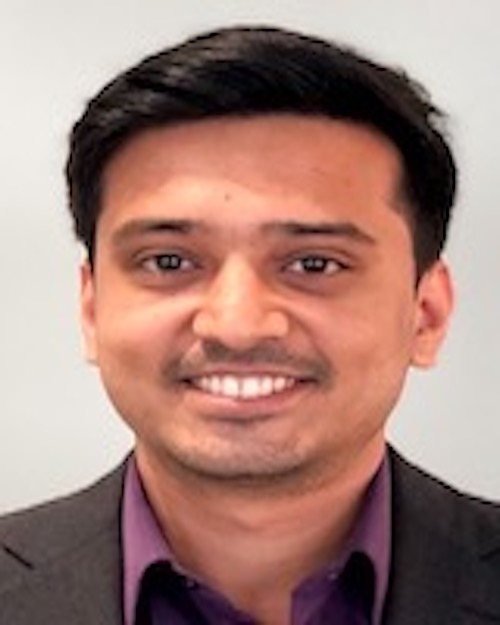}}]
{Prabhakar Ray} (Student Member, IEEE) received the B.E. degree in electrical and electronics engineering from PES University, Bengaluru, India, in 2013, and the M.Sc. degree in web intelligence from King's College London, London, U.K., in 2016.  He is currently working towards the Ph.D. degree, focusing on robotic untangling of tangled granular materials with the Robot Learning Lab (RLL) at King's College London, London, U.K.
\end{IEEEbiography}

\begin{IEEEbiography}[{\includegraphics[width=1in,height=1.25in,clip,keepaspectratio]{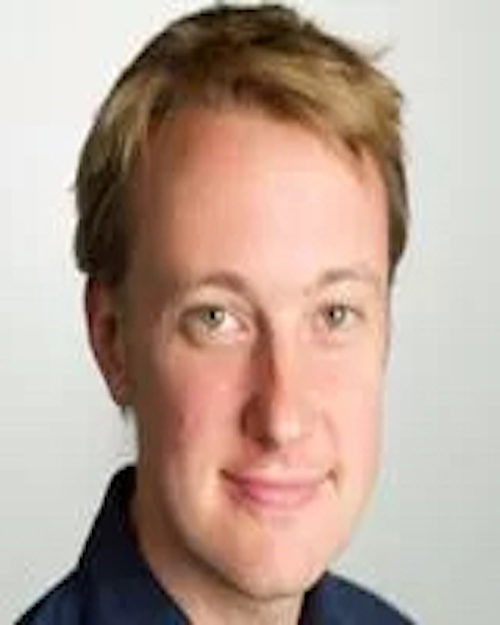}}]
{Dr Matthew Howard} (Member, IEEE) is Reader at the Centre for Robotics Research, Department of Engineering, King's College London. Prior to joining King's in 2013, he held a Japan Society for Promotion of Science fellowship at the Department of Mechanoinformatics at the University of Tokyo and was a research fellow at the University of Edinburgh from 2009-2012. He also obtained his PhD in 2009 at Edinburgh with an EPSRC CASE award sponsored by Honda Research. His research interests span the fields of robotics and autonomous systems, statistical machine learning and adaptive control. His current work focuses on teaching and learning of robotic motor skills by demonstration, especially for soft robotic devices, based on human musculoskeletal control. He works with a number of large companies in bringing automation to agri-food production through collaborative robots.
\end{IEEEbiography}








\end{document}

%% file: tex/preamble.tex
\IEEEoverridecommandlockouts                              

\usepackage{times}
\usepackage{graphicx}\graphicspath{{img/}{images/}}
\usepackage{epstopdf}
\usepackage{algorithm}
\usepackage{algorithmic}
\usepackage{multirow} 
\usepackage{multicol}
\usepackage{array}
\usepackage{mdwmath}
\usepackage{mdwtab}
\usepackage{rotating}
\usepackage[list=off]{caption}
\usepackage{amssymb}
\usepackage{verbatim}
\usepackage{overpic}
\usepackage{amsmath}
\usepackage{subcaption}
\usepackage[utf8]{inputenc}
\usepackage[export]{adjustbox}
\usepackage[hidelinks]{hyperref}
\usepackage[normalem]{ulem}
\usepackage[background = white, text=black , arrow = black, line width=5mm]{callouts}

\makeatletter\newcommand{\manuallabel}[2]{\def\@currentlabel{#2}\label{#1}}\makeatother

\usepackage[
backend=bibtex,
style=ieee,
url=false,
doi=false,
isbn=false,
doi=true,
eprint=false,
mincitenames=1,
maxcitenames=2
]{biblatex}
\input{tex/abbreviations}
\input{tex/symbols}

\input{tex/format}

\input{tex/comments}

%% file: tex/abbreviations.tex
\input{ext/tex/abbreviations}

\renewcommand  {\DoF}{degree of freedom\xspace}
\providecommand{\GI} {graspability index\xspace}
\providecommand{\SP}{Spread-and-Pick\xspace}
\providecommand{\FP}{Fixed-Point-Picking\xspace}

\newacronym{CNN}{CNN}{convolution neural network}

\newacronym{MAE}{MAE}{Mean Absolute Error}

\newglossaryentry{GM} 
{
	name={GM},
	description={granular material},
	first={\glsentrydesc{GM} (\glsentrytext{GM})},
	plural={GMs},
	descriptionplural={granular materials},
	firstplural={\glsentrydescplural{GM} (\glsentryplural{GM})}
}\providecommand{\GM}{\gls{GM}\xspace}\providecommand{\GMs}{\glspl{GM}\xspace}
\newacronym{SnP}{SnP}{spread-and-pick}\renewcommand{\SP}{\gls{SnP}\xspace}
\newacronym{DoF}{DoF}{degree of freedom}\renewcommand{\DoF}{\gls{DoF}\xspace}
\newacronym{FP}{FP}{fixed-point-picking}\renewcommand{\FP}{\gls{FP}\xspace}
\newacronym{GI}{GI}{graspability index}\renewcommand{\GI}{\gls{GI}\xspace}
\newglossaryentry{PL} 
{
	name={PL},
	description={protrusion length},
	first={\glsentrydesc{PL} (\glsentrytext{PL})},
	plural={\glsentrytext{PL}s},
	descriptionplural={\glsentrydesc{PL}s},
	firstplural={\glsentrydescplural{PL} (\glsentryplural{PL})}
}\providecommand{\PL}{\gls{PL}\xspace}\providecommand{\PLs}{\glspl{PL}\xspace}

%% file: ext/tex/abbreviations.tex
\usepackage{xspace}
\usepackage[acronym]{glossaries}%
\glsdisablehyper 

\providecommand{\DoF    }{degrees of freedom\xspace}



%% file: tex/symbols.tex
\input{ext/tex/symbols}

\usepackage{siunitx} 

\providecommand{\conv}{*} 

\providecommand{\mt}     {m_t}                  

\providecommand{\r}      {r}                  %
\renewcommand  {\r}      {r}                  %
\providecommand{\br}     {\mathbf{\r}}        %
\providecommand{\rx}     {\r_x}               
\providecommand{\ry}     {\r_y}               
\providecommand{\rz}     {\r_z}               
\providecommand{\rt}     {\r_{\theta}}  

\providecommand{\Or}     {\mathbf{u}} 
\providecommand{\Er}     {\mathbf{v}} 
\providecommand{\Oc}     {\mathbf{O_c}}         
\providecommand{\Ocp}     {\mathbf{O_{c'}}}          

\providecommand{\gc}     {\mathbf{G_c}}   
\providecommand{\gcp}     {\mathbf{G_{c'}}}   

\providecommand{\wc}     {\mathbf{W_c}}   
\providecommand{\wcp}     {\mathbf{W_{c'}}}   

\providecommand{\g}       {\mathbf{g}}  
\providecommand{\G}       {\mathbf{G}} 
\providecommand{\w}       {w} 
\providecommand{\p}       {p} 
\providecommand{\l}       {l} 

\providecommand{\pickpar}   {\delta}

\providecommand{\e}       {PE} 
\providecommand{\m}       {m} 
\providecommand{\f}       {f} 

%% file: ext/tex/symbols.tex
\usepackage{amssymb}
\usepackage{mathtools}
\usepackage{amsmath}
\usepackage{gensymb}

\mathchardef\mhyphen="2D   

\providecommand{\R}     {\mathbb{R}}          
\providecommand{\T}     {\top}                

\providecommand{\br}     {\mathbf{r}}         

\providecommand  {\nu}  {q}                   



%% file: tex/format.tex
\input{ext/tex/format}

\usepackage{siunitx}

\renewcommand{\tref}[1]{Table~\ref{t:#1}}   

\providecommand{\cl}[1]{\begin{enumerate*}[label=(\alph*)]#1\end{enumerate*}}

%% file: ext/tex/format.tex
\usepackage{xspace}

\providecommand{\eg}{\textit{e.g.,}~} %
\providecommand{\ie}{\textit{i.e.,}~} %
\providecommand{\figurename}{Fig.}%
\providecommand{\tablename}{Table}%
\providecommand*{\sref}[1]{\S\ref{s:#1}}            
\providecommand*{\tref}[1]{\tablename~\ref{t:#1}}   
\providecommand*{\fref}[1]{\figurename~\ref{f:#1}}  
\providecommand*{\eref}[1]{(\ref{e:#1})}            

\usepackage[inline]{enumitem}
\setlist{nolistsep}
\providecommand{\il}[1]{\begin{enumerate*}[label=(\roman*)]#1\end{enumerate*}}

%% file: tex/comments.tex
\usepackage{ifdraft}
\usepackage[normalem]{ulem}                                        
\usepackage{marginnote}
\usepackage[textwidth=\marginparwidth,colorinlistoftodos]{todonotes}

\colorlet{jb}{red}
\colorlet{mh}{red}
\colorlet{pr}{green}

\providecommand  {\note}[1]{{\color{gray}#1}}

\providecommand  {\mhsout}{\bgroup\markoverwith{\textcolor{mh}{\rule[0.5ex]{2pt}{0.4pt}}}\ULon} 